\pdfoutput=1

\documentclass[11pt]{article}

\usepackage{latexsym}
\usepackage{multirow}
\usepackage{inconsolata}
\usepackage{mathrsfs}

\usepackage{times}
\usepackage[utf8]{inputenc} %
\usepackage[T1]{fontenc}    %
\usepackage[hidelinks]{hyperref}       %
\usepackage{natbib}
\usepackage{url}            %
\usepackage{booktabs}       %
\usepackage{amsfonts}       %
\usepackage{nicefrac}       %
\usepackage{microtype}      %
\usepackage{enumitem}
\usepackage{bbm}

\usepackage{graphicx}
\usepackage{sidecap}
\usepackage{diagbox}
\usepackage[small]{caption}
\usepackage{subcaption}
\usepackage{comment} 
\usepackage{array}
\usepackage[export]{adjustbox}

\usepackage{float} %
\usepackage{wrapfig}
\usepackage{mathtools}
\usepackage{xspace}
\usepackage{bm} %

\newcommand{\rightcomment}[1]{\(\triangleright\) {\small \it #1}}  %
\newcommand{\eqcomment}[1]{\addtocounter{equation}{1}\tag*{\rightcomment{#1}\quad(\theequation)}}  %
\usepackage{suffix}
\WithSuffix\newcommand\eqcomment*[1]{\tag*{\rightcomment{#1}}}  %

\usepackage{algorithm}
\usepackage[noend]{algpseudocode}

\renewcommand\algorithmicthen{:}
\algnewcommand{\IfThen}[2]{\State \algorithmicif\ #1\ \algorithmicthen\ #2}
\algnewcommand{\IfThenElse}[3]{\State \algorithmicif\ #1\ \algorithmicthen\ #2\ \algorithmicelse\ #3}
\algrenewcommand{\algorithmiccomment}[1]{\hfill \rightcomment{#1}}
\algnewcommand{\LineComment}[1]{\State \rightcomment{#1}}
\algnewcommand{\LinesComment}[1]{\State \rightcomment{\parbox[t]{\linewidth-\leftmargin-\widthof{\(\triangleright\) }}{#1}}\smallskip}

\algnewcommand\algorithmicinput{{\bfseries Input:}}
\algnewcommand\INPUT{\item[\algorithmicinput]}
\algnewcommand\algorithmicoutput{{\bfseries Output:}}
\algnewcommand\OUTPUT{\item[\algorithmicoutput]}
\makeatletter
\newcounter{algorithmicH}
\let\oldalgorithmic\algorithmic
\renewcommand{\algorithmic}{%
  \stepcounter{algorithmicH}
  \oldalgorithmic}
\renewcommand{\theHALG@line}{ALG@line.\thealgorithmicH.\arabic{ALG@line}}
\makeatother
\makeatletter
\newcommand{\algmargin}{\the\ALG@thistlm}
\makeatother
\algnewcommand{\Statepar}[1]{\State\parbox[t]{\dimexpr\linewidth-\algmargin}{\strut #1\strut}}

\usepackage{mdframed} %

\usepackage[usenames,dvipsnames]{xcolor}
\usepackage[textwidth=1.5cm]{todonotes} %
\usepackage{pgfplots}

\newcommand{\para}[1]{\noindent \textbf{#1}}
\newcommand{\cutforspace}[1]{}

\usepackage{listings}

\usepackage{parcolumns}
\lstdefinestyle{datalogstyle}{
        basicstyle={\tt \scriptsize},  %
	xleftmargin={6pt},
        xrightmargin={6pt},
        columns=flexible,
        breakindent=0pt,
        breaklines=true, 
	frame=tb,
	stepnumber=1,
	firstnumber=1,
	numberfirstline=true,
	tabsize=2,
	extendedchars=true,
	breaklines=true,
	columns=fullflexible,
	keepspaces=true,
	escapeinside={@}{@},
	firstnumber=last,
	captionpos=b, 
	commentstyle=\color{black!65},
	numberstyle=\tiny\color{black!65},
	stringstyle=\color{codepurple},
	breakatwhitespace=false, 
	keepspaces=true,              
        mathescape=true, 
	numbersep=5pt,                  
	showspaces=false,                
	showstringspaces=false,
	showtabs=false,
	aboveskip={0.8\baselineskip},
	belowskip={0.2\baselineskip},
}
\definecolor{aigold}{RGB}{244,210, 1} 
\definecolor{aigreen}{RGB}{213, 245, 227}

\definecolor{humanpurple}{RGB}{235, 222, 240} 
\definecolor{mypurple}{RGB}{147,112,219} 
\definecolor{myorange}{RGB}{255,165,0} 
\definecolor{commentgray}{RGB}{86, 101, 115}
\definecolor{mygray}{RGB}{169,169,169}
\definecolor{aired}{RGB}{255,180,181}

\usepackage[final]{acl}

\usepackage{graphicx}

\newif\ifhidecomments

\hidecommentstrue

\ifhidecomments
    \newcommand{\chenhao}[1]{}
    \newcommand{\haokun}[1]{}
    \newcommand{\rosa}[1]{}
    \newcommand{\aldous}[1]{}
    \newcommand{\chenfei}[1]{}
\else
    \newcommand{\chenhao}[1]{\textcolor{blue}{[\textsc{Chenhao}: #1]}}
    \newcommand{\haokun}[1]{\textcolor{green!30!brown}{[\textsc{Haokun}: #1]}}
    \newcommand{\rosa}[1]{\textcolor{magenta!80!brown}{[\textsc{Rosa}: #1]}}
    \newcommand{\aldous}[1]{\textcolor{yellow!10!brown}{[\textsc{Aldous}: #1]}}
    \newcommand{\chenfei}[1]{\textcolor{pink}{[\textsc{Chenfei}: #1]}}
\fi

\usepackage[noabbrev,capitalize]{cleveref} %
\crefname{equation}{eq.}{eqs}   %
\crefname{footnote}{footnote}{footnotes}   
\crefname{listing}{Example}{Examples}
\crefname{assumption}{assumption}{assumptions}
\crefname{line}{line}{lines}   %
\crefname{section}{\S}{\S\S}
\creflabelformat{equation}{#2#1#3}

\newcommand{\calS}{\mathcal{S}}
\newcommand{\calH}{\mathcal{H}}
\newcommand{\calM}{\mathcal{M}}
\newcommand{\calL}{\mathcal{L}}
\newcommand{\calW}{\mathcal{W}}
\newcommand{\calP}{\mathcal{P}}
\newcommand{\calD}{\mathcal{D}}
\newcommand{\calC}{\mathcal{C}}

\newcommand{\deceptive}{\textsc{Deceptive reviews}\xspace}
\newcommand{\llamagc}{\textsc{LlamaGC}\xspace}
\newcommand{\gptgc}{\textsc{GPTGC}\xspace}
\newcommand{\persuasion}{\textsc{Persuasive Pairs}\xspace}
\newcommand{\dreaddit}{\textsc{Dreaddit}\xspace}
\newcommand{\fourcities}{\textsc{Four-cities}\xspace}
\newcommand{\writingprompts}{\textsc{WritingPrompts}\xspace}

\newcommand{\llama}{\textsc{Llama-3.1-70B-Instruct}\xspace}
\newcommand{\gpt}{\textsc{GPT-4o-mini}\xspace}
\newcommand{\shortllama}{\textsc{Llama-70B-I}\xspace}
\newcommand{\shortgpt}{\textsc{GPT-4-mini}\xspace}

\newcommand{\notebooklm}{\textsc{NotebookLM}\xspace}
\newcommand{\hyperwrite}{\textsc{HyperWrite}\xspace}
\newcommand{\hypogenic}{\textsc{HypoGeniC}\xspace}
\newcommand{\paperonly}{\textsc{literature-only}\xspace}
\newcommand{\refinemethod}{\textsc{HypoRefine}\xspace}

\newcommand{\unionhypogenic}{\textsc{Literature}$\cup$\hypogenic\xspace}
\newcommand{\unionhyporefine}{\textsc{Literature}$\cup$\refinemethod\xspace}

\definecolor{pastelgreen}{RGB}{50,205,50}
\newcommand{\increase}{\textcolor{pastelgreen}{\bm{$\uparrow$}}}
\newcommand{\decrease}{\textcolor{red}{\bm{$\downarrow$}}}

\title{Literature Meets Data: A Synergistic Approach to Hypothesis Generation}

\author{Haokun Liu$^{\clubsuit *}$, Yangqiaoyu Zhou$^{\clubsuit *}$, Mingxuan Li$^{\clubsuit *}$, Chenfei Yuan$^\dag$\\
 \& \textbf{Chenhao Tan}$^\clubsuit$  \\
Department of Computer Science\\
University of Chicago$^\clubsuit$, Tsinghua University$^\dag$\\
Chicago, IL 60637, USA \\
\texttt{\{haokunliu, zhouy1, mingxuanl, chenhao\}@uchicago.edu}, \\ 
\texttt{yuancf21@mails.tsinghua.edu.cn} \\
}

\begin{document}

\maketitle

\renewcommand{\thefootnote}{\fnsymbol{footnote}}
\footnotetext[1]{Equal contributions.}

\begin{abstract}
AI holds promise for transforming scientific processes, including hypothesis generation.
Prior work on hypothesis generation can be broadly categorized into theory-driven and data-driven approaches. 
While both have proven effective in generating novel and plausible 
hypotheses, 
it remains an open question whether they can complement each other.
To address this, we develop the first method that combines literature-based insights with data to perform LLM-powered hypothesis generation. We apply our method on five different 
datasets and demonstrate that integrating literature and data outperforms other baselines (8.97\% over few-shot, 15.75\% over literature-based alone, and 3.37\% over data-driven alone). 
Additionally, we conduct the first human evaluation to assess the utility of LLM-generated hypotheses in assisting human decision-making on two challenging tasks: deception detection and AI generated content detection.
Our results show that human accuracy improves significantly by 7.44\% and 14.19\% on these tasks, respectively.
These findings suggest that integrating literature-based and data-driven approaches provides a comprehensive and nuanced framework for hypothesis generation and could open new avenues for scientific inquiry.
\end{abstract}

\section{Introduction}
\label{sec:introduction}

\begin{figure*}[t]
    \centering
    \includegraphics[trim={0cm 0cm 0cm 0.4cm}, clip, width=0.9\textwidth]{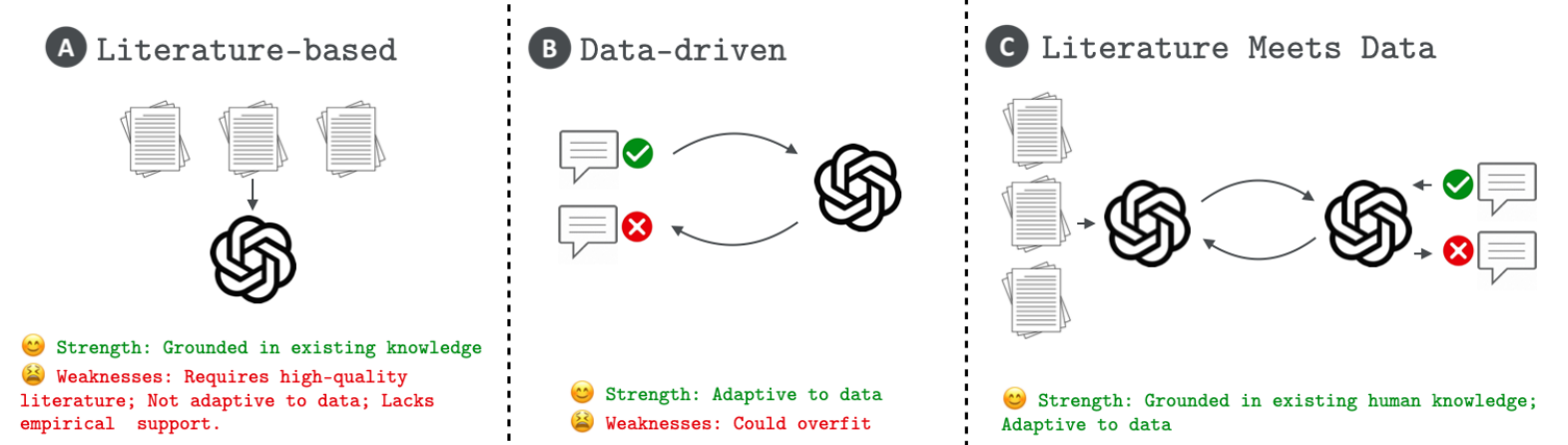}
    \caption{
    The literature-based approach (A) leverages existing human knowledge to generate hypotheses but struggles to adapt to new data and lacks empirical grounding. The data-driven approach (B) relies on large datasets to generate hypotheses, enabling adaptation to diverse scenarios but risking overfitting. The literature + data approach (C) combines the strengths of both, grounding hypotheses in human knowledge while incorporating empirical data to enhance adaptiveness.
    See algorithmic details in \cref{sec:methods}.
    }
    \label{fig:illustration}
\end{figure*}

\begin{quote}
``It is the theory that decides what can be observed.''
\hfill ---Albert Einstein
\end{quote}

Large language models (LLMs) excel at synthesizing information and hold promise for transforming hypothesis generation, a critical yet understudied step in scientific discoveries.
Many recent studies recognize this potential and use LLMs to generate hypotheses~\cite[e.g.,][]{yang2024moose, batista2024words}.
We broadly categorize them into \textit{theory-driven} and \textit{data-driven} methods. 

On one hand, theory-driven approaches leverage LLMs to review existing literature and generate novel hypotheses/ideas \citep{yang2024moose, baek2024researchagent}.
These methods have shown promising results in terms of the hypotheses' novelty, validity, and usefulness to researchers, while remaining grounded in established human knowledge \citep{si2024llmsgeneratenovelresearch}.
However, they come with notable limitations: they require high-quality literature, struggle to adapt to new data, and lack empirical support.
Data-driven approaches, on the other hand, propose hypotheses by discovering patterns in data \cite{zhou2024hypogenic, qiu2024phenomenal}.
These hypotheses are data-adaptive and can exhibit strong performance in explaining the data.
However, they could be too overly tailored to the specific datasets used, which can hinder their generalizability.

We hypothesize that theory can guide the discovery from data
and propose 
to integrate literature-based and data-driven hypothesis generation (see \cref{fig:illustration}).
For the data-driven component, we use \hypogenic as the backbone \cite{zhou2024hypogenic}. 
\hypogenic leverages an LLM to initialize hypotheses from a small number of examples and then updates them iteratively to improve the quality of hypotheses. 
To enhance this process with literature insights, we introduce a literature-based hypothesis agent.
This agent interacts with the data-driven hypothesis agent (\hypogenic), refining and maintaining a shared pool of hypotheses through continuous collaboration, ensuring that the hypotheses benefit from both data-driven adaptability and the grounding of existing scientific knowledge.
In addition to the refinement approach, we also propose to directly unionize literature-based and data-driven hypotheses.

To comprehensively evaluate these hypotheses, we conduct automatic and human evaluation to assess their generalizability, utility, and novelty.
We apply our method to address research questions in social sciences: deception detection, AI generated content (AIGC) detection, mental stress detection, and persuasive argument prediction.
Automatic evaluation results show that integrating literature and data outperforms other baselines: 8.97\% over few-shot, 15.75\% over literature-based alone, and 3.37\% over data-driven alone in accuracy on out-of-distribution datasets, a measure of generalizability.

Moreover, we conduct the first study to assess the utility of AI-generated hypotheses in improving human decision-making and show 
that our generated hypotheses improve human accuracy by 7.44\% and 14.19\% on deception detection and AIGC detection. 
Additionally, we find that literature-based and data-driven hypotheses complement each other, as one set often contains novel information not found in the other set.

In sum, we make the following contributions:
\begin{itemize}[leftmargin=*,itemsep=0pt,topsep=-2pt]
\item We propose the first approach to using both literature information and data for LLM-powered hypothesis generation.

\item We conduct automatic evaluation to assess the utility of the generated hypotheses in improving LLM predictions.
Experiments on five datasets demonstrate the effectiveness of our approach.

\item We conduct the first human evaluation to test the utility of LLM-generated hypotheses and demonstrate consistent improvements on two challenging tasks.
\end{itemize}

\noindent The code and data are available at \url{https://github.com/ChicagoHAI/hypothesis-generation}.

\section{Methods}
\label{sec:methods}

We formulate the problem of hypothesis generation as follows.
Assuming that we have access to literature $\calL$ and observational data $\calD$ that are relevant to a research question $q$. Then, we want to develop an AI-powered algorithm $f$ with model $\calM$ such that we can generate high-quality hypotheses for the research question $q$, i.e., $\calH = f_{\calM}(q, \calL,\calD)$.
Example research questions include what makes an argument persuasive and what signs are indicative of AI-generated texts.
In this work, we consider research questions that can be formulated as classification tasks, so we use $q$ and \textit{task} interchangeably.

\subsection{Literature-Based Hypothesis Generation}
\label{sec:literature-based-hypothesis-generation}

For the \paperonly method, we start by picking a set of papers $\calP=\{p_1, p_2, ..., p_m\}$ for $q$ from related papers 
on Semantic Scholar or Google Scholar. We also choose from papers that cited the original datasets for each task. Subsequently, we use S2ORC-doc2json  
to convert the raw PDF versions of the papers to a corpus of JSON files \citep{lo-wang-2020-s2orc}. We denote the converted papers as $\calC=\{\text{doc2json}(p):p\in\calP\}$. Passing the full texts of all these papers  to a language model would likely exceed its maximum context length. Moreover, we want to generate hypotheses from the key findings of the relevant papers because some contents, such as technical details, may not help significantly but distract the LLM. 
Therefore, we develop a paper summarizer $\calM_S$ to generate paper summaries $\calS=\{\calM_S(p_c):p_c\in\calC\}$ (throughout the paper, we use subscripts
to indicate $\calM$ with different prompts).
Lastly, we instruct language models to generate hypotheses $\calH_\calL=\calM_{G}(\calS)$ based on the generated paper summaries, with an emphasis on usefulness for carrying out the specific tasks that our literature corpus focuses on. 

In addition to our own implementation, we use commercial ones such as \notebooklm \citep{google2024notebooklm} and \hyperwrite \citep{othersideai2024hyperwrite} as strong baselines.

\subsection{Data-Driven Hypothesis Generation}
\label{sec:data-driven-hypothesis-generation}

Our data-driven hypothesis generation adopts \hypogenic%
in \citet{zhou2024hypogenic}.
Here we give a brief overview.
Suppose we have a set of observational data in the form of input-label pairs, i.e., $\calD=\{(x_1,y_1),...,(x_n,y_n)\}$. During the initialization stage of \hypogenic, a generation agent $\calM_G$ is prompted with a set of initial data instances $\calD_{\operatorname{init}} \subset \calD$ and asked to generate initial hypotheses $\calH_{\calD}^0=\calM_G(\calD_{\operatorname{init}})$. Then, for each of the initial hypothesis $h$ and for every example $(x_i,y_i)\in \calD_{\operatorname{init}}$, $h$ is used to prompt an inference agent $\calM_I$ to make a prediction $\hat{y}_{i} = \calM_I(x_{i}, h)$. The initial reward of each hypothesis is computed using:
\begin{equation}
\label{eq:reward_function}
\begin{aligned}
    r_t(h) & \coloneqq \operatorname{Acc}(h,\mathbf{X}_h^t) + \alpha\sqrt{\frac{\log{t}}{|\mathbf{X}_h^t|}}, \\
    \operatorname{Acc}(h,\mathbf{X}) & \coloneqq \frac{\sum_{(x_i, y_i) \in \mathbf{X}} \mathbbm{1}(y_i = \calM_I(x_i, h))}{|\mathbf{X}|},
\end{aligned}
\end{equation}
where $r_t$ is the reward function inspired by the upper confidence bound (UCB) algorithm \citep{Auer2003UsingCB}, $\mathbf{X}_h^t$ is the set of examples being used to evaluate hypothesis $h$ up to time $t$, 
and $\alpha$ is the reward coefficient that controls the exploration term of the reward function.
We also initialize $\calW=\emptyset$, where $\calW$ keeps track of the wrongly predicted examples. 

In the update stage at time step $t+1$, we take $(x_{t+1},y_{t+1}) \in \calD$, and the top $k$ high-reward hypotheses.
For each $h$ of the selected hypotheses, we prompt $\calM_I$ to make a prediction $\calM_I(x_{t+1},h)$. The accuracy and reward of the $k$ hypotheses are updated using \Cref{eq:reward_function}.
Among the $k$ hypotheses, if at least $w_{\operatorname{hyp}}$ predicted wrong, We add the example to $\calW$. If $|\calW|\geq w_{\operatorname{max}}$, a set of new hypotheses $\calH_{\calW}=\calM_G(\calW)$ is generated and added to $\calH_\calD^{t+1}$. Then the top $H_{\operatorname{max}}$ hypotheses are kept, and $\calW$ is reset to $\emptyset$. After all $n$ examples are visited, we denote the final hypothesis bank as $\calH_\calD$.

\subsection{Integration of Literature-based and Data-driven Hypotheses}
\label{sec:integration-of-literature-based-and-data-driven-hypotheses}

One main contribution of our work is proposing the first approach to integrate literature-based and data-driven hypothesis generation so that we can effectively leverage the strengths of each approach, increasing the generalizability and utility of generated hypotheses.
We consider two strategies.

\paragraph{Refinement of literature-based hypotheses.}
\label{sec:refinement-of-literature-based-hypotheses}
\refinemethod integrates paper summaries $\calS$ from \cref{sec:literature-based-hypothesis-generation} with \hypogenic. In the initialization stage, a generation agent $\calM_G$ is asked to generate initial hypotheses based on both a set of initial examples and paper summaries, given by $\calH_{\calL+\calD}^0=\calM_G(\calS, \calD_{\operatorname{init}})$. 

In the update stage, we propose an iterative refinement approach to integrate patterns from data and key findings from literature into new hypotheses. Specifically, each time \hypogenic generates a set of new hypotheses $\calH_\calW$, these hypotheses are refined multiple rounds by a data-driven refinement agent and a literature-based refinement agent. Take $\calM_R$ as the refinement agent, each time $\calH_\calW$ is generated from the wrong examples pool $\calW$, it is iteratively refined as follows: 
\begin{equation*}
    \calH_\calW^{i} = \begin{cases}
    \calM_R(\calH_\calW^{i-1}, \calS) & \text{if } i \bmod 2 = 0 \\
    \calM_R(\calH_\calW^{i-1}, \calW) & \text{if } i \bmod 2 = 1.
\end{cases}
\end{equation*}
After $N_{\operatorname{refine}}$ rounds of refinement, the final hypothesis bank $\calH_\calW^{N_{\operatorname{refine}}}$ is fed back to the \hypogenic pipeline as $\calH_\calW$. 

The reward function and update process for the hypothesis bank $\calH_{\calL+\calD}^t$ remain consistent with those of the original \hypogenic. 

\paragraph{Union and redundancy elimination.}
\label{sec:union-and-deduplication}
As the reward function of \hypogenic focuses only on the hypotheses' performance on the datasets at hand, literature-based hypotheses are sometimes undervalued during the update stage. On occasions they can even be replaced by hypotheses that have especially good performances on data but are not necessarily generalizable on real-world tasks. To counter this issue, we use a union approach to combine literature-based and data-based hypotheses. We first generate two hypothesis banks: one literature-based hypothesis bank $\calH_\calL$ and the other bank from $\calH_\calD$ or $\calH_{\calL+\calD}$, using \hypogenic or \refinemethod, respectively.
Then we build a redundancy checker to remove hypotheses that express overly similar or repeating information in each bank. Lastly, we construct the final hypothesis bank of size $H_{\operatorname{max}}$ by randomly 
choosing $\frac{H_{\operatorname{max}}}{2}$ hypotheses from the literature-based hypothesis bank and adding the top $\frac{H_{\operatorname{max}}}{2}$ hypotheses from the other hypothesis bank based on training accuracies. For detailed information of the implementation,
please refer to \cref{appendix:implementation_details:refine_union_implementation}.

\section{Experiments}
\label{sec:experiments}

In this section, we introduce our evaluation framework and the tasks to operationalize it.

\subsection{Evaluation Framework}
\label{sec:evaluation}
Formally evaluating hypotheses requires rigorous 
protocols and vast amounts of resources. In this work, we mainly evaluate our generated hypotheses along two dimensions: utility and novelty. We perform both automatic and human evaluations to show that our generated hypotheses can help models and humans in challenging real-world classification tasks and bring novel information.

\paragraph{Automatic evaluation on out-of-distribution (OOD) and in-distribution (IND) datasets and cross-model inference.}
Since we work with classification tasks, a natural way of evaluating the hypotheses is prompting the LLMs to do inference with the hypotheses. For all methods that generate hypotheses $\calH$, with every test example $(x,y)$, we prompt $\calM_I$ to first extract the most relevant hypotheses to the example and make inference using the hypotheses, denoted as $\calM_I(\calH, x)$. For detailed information about the prompts, see \cref{appendix:prompts}. Then we compute $\operatorname{Acc}(\calH,\calD_{\operatorname{test}})$, defined in \cref{eq:reward_function}, for a held-out set $\calD_{\operatorname{test}}$. For each task, we report average accuracy and F1 scores on held-out OOD and IND sets for 5 different random seeds. 
Since we are most interested in the generalizability of the generated hypotheses, we focus on performance on the OOD set in the main paper.

In addition to predicting on out-of-distribution datasets, 
we test our hypotheses' generalizability by taking the hypotheses generated by one model and performing
inference with another model.

\paragraph{Human evaluation on utility and novelty.}
We design human studies to assess the practical utility of the generated hypotheses 
on Deception Detection and AIGC Detection. 
In addition, we evaluate the perceived clarity, novelty, and plausibility through surveys.
Screenshots and details of the studies are in \cref{appendix:human_study}. We pay participants at an average hourly rate of \$12.

\para{Human Study I: Utility in human decision-making.} We recruit 60 Prolific participants and randomly assign them into experimental and control groups. The control group performs the task without hypotheses, while the experiment group is given a set of three generated hypotheses to complete the same task. 
Specifically, each participant is randomly assigned 14 instances, and we include attention check questions to ensure the quality of the collected responses.
We evaluate the practical utility of our generated hypotheses by comparing the performance of the two groups. 

We pick the hypotheses based on their impact on performance in an ablation setting. Specifically, we choose the top three hypotheses that cause the greatest drop in performance when removed from the hypotheses pool during multi-hypothesis inference. In addition, to motivate participants to perform at their best, we offer a bonus of \$0.1 for each correctly predicted instance.

At the end of the study, participants in the experiment group are also asked to give overall ratings and an assessment of the given hypotheses. There are five scales: ``Not at all helpful'', ``Slightly helpful'', ``Moderately helpful'', ``Very helpful'', and ``Extremely helpful''. 

\paragraph{Human study II: Clarity, novelty, and plausibility.} 
We recruit 30 participants with graduate-level degrees in social sciences from \textit{prolific.com} to evaluate hypotheses generated by \refinemethod, \notebooklm, and \hyperwrite for Deception Detection. Using a 5-point Likert scale, participants assess each hypothesis along three dimensions: clarity, novelty, and plausibility.
See details in \cref{appendix:human_study:likert}.

\paragraph{Human study III: Novelty and nuance.} 
To compare data-driven hypotheses and literature-driven hypotheses, we present one hypothesis of each type to participants and ask them to judge whether the second hypothesis provides meaningfully novel information that is not covered in the first hypothesis.

We sample 50 pairs of hypotheses $(h_1, h_2)$, one from literature-based and one from data-driven, with duplications removed within each group.
We recruit 10 Prolific participants to annotate whether $h_2$ provides new information to $h_1$ for each pair. Each participant is randomly assigned to annotate 15 pairs. For each pair, we take the majority vote to determine the final novelty label.

\subsection{Tasks}
\label{sec:tasks}
We consider four tasks in social sciences. 

\noindent \textbf{Deception Detection} is a 
widely studied problem in psychology and other social sciences \citep{granhag2005deception}. 
We use the dataset introduced by \citet{ott-etal-2013-negative} (\deceptive), which consists of 800 genuine hotel reviews and 800 fake hotel reviews, as our IND dataset. 
For the OOD dataset, we use hotel reviews from different source websites and different cities \citep{oodhotelreviews}.

\noindent \textbf{AI-Generated Content (AIGC) Detection} 
has attracted significant attention in recent years \citep{tang2023sciencedetectingllmgeneratedtexts}. 
Most existing works focus on developing black-box detection methods and rarely take interpretability into account \citep{wu2024surveyllmgeneratedtextdetection}. 
We thus build our own dataset for this task. We take 800 distinct prompts and human-written stories in the \writingprompts dataset \citep{fan-etal-2018-hierarchical}. Then we use the same prompts to generate AI-written stories with \textsc{Llama-3.1-70B-Instruct} \citep{dubey2024llama3herdmodels} and \textsc{GPT-4o-mini} \citep{openai2023gpt4}, constituting our \llamagc and \gptgc datasets. The IND data contains stories generated by the corresponding model. The stories generated by the other model are treated as OOD data.

\noindent \textbf{Persuasive Argument Prediction} examines persuasion and social interactions to reveal predictive cues of persuasiveness \citep{Tan_2016}. We use \persuasion, a dataset with pairs of short texts constructed by \citet{pauli2024measuringbenchmarkinglargelanguage}. Within each pair of texts, one is from existing corpora with signals of persuasiveness, while the other one is generated by an LLM with instructions for it to be more/less persuasive than the one from existing corpora. 
We formulate this task as predicting the more persuasive one of each pair of texts. The dataset contains human-annotated ground-truth labels and is pre-processed by removing examples where there exists disagreement among annotators.
The IND and OOD datasets are then created based on different original sources of texts.

\noindent \textbf{Mental Stress Detection} from social media content is 
an important task in mental health~\citep{lupien_stress_review}. 
We use \dreaddit, a corpus of lengthy Reddit posts with stress status labels developed by \citet{turcan2019dreadditredditdatasetstress}. The dataset contains 3.5k post segments annotated using Amazon Mechanical Turk, with labels indicating the presence or absence of stress in posts. Our IND and OOD sets are separated based on 
subreddits that the posts come from.

For each task, we split the IND dataset with at least 200 examples in train set, 300 in test set (on which we perform inference), 300 in validation set, 
and sample at least 300 instances from OOD (see \cref{appendix:implementation_details:OOD_parititioning} for more details). %

\subsection{Implementation and Baselines}
\label{sec:baseline}

Our method works with any LLM ($\mathcal{M}$). 
We use \gpt and \llama in the main paper.
We refer to \gpt as ``\shortgpt'' and \llama as ``\shortllama''.
We compare our method with the following baselines.

\begin{enumerate}[leftmargin=*,itemsep=0pt,topsep=-2pt]
    \item {\bf Zero-shot and few-shot prompting.} 
    We give the LLMs detailed task instructions (zero-shot) and optionally provide three demonstrating examples (few-shot).
    This approach does not involve any hypothesis.

    \item {\bf Zero-shot hypothesis generation.} Inspired by \citet{qi2023large}, we provide specific task descriptions and instructions, and then we prompt the LLMs to generate hypotheses directly without incorporating literature or data.

    \item {\bf Literature-driven hypothesis generation.} We use the implementation in \S\ref{sec:literature-based-hypothesis-generation}. 
    In addition to our own implementation, we compare two of the recently released agent frameworks for scientific writing, \notebooklm \citep{google2024notebooklm} and \hyperwrite \citep{othersideai2024hyperwrite}. We use the same prompt for \notebooklm and \hyperwrite as what we apply in our methods. See details in \cref{appendix:implementation_details:notebooklm_hyperwrite}.
    These methods only use literature in hypothesis generation.

    \item {\bf Data-driven hypothesis generation}. We use \hypogenic. See details in \cref{sec:data-driven-hypothesis-generation}.
\end{enumerate}

For all the hypothesis generation methods we use, we keep the size of the hypothesis bank $\calH$ to be 20 (i.e., $H_{\operatorname{max}}=20$.)

\section{Results}
\label{sec:results}
\begin{table*}[t]
\centering
\resizebox{0.95\textwidth}{!}{%
\begin{tabular}{@{}ll@{}rrrrr@{}}
\toprule
Model & Methods & \deceptive & \llamagc & \gptgc & \persuasion & \dreaddit \\

\midrule

\multirow{13}{*}{\shortstack{\textsc{GPT-4} \\ \textsc{MINI}}} 
    & \multicolumn{6}{c}{\textbf{No hypothesis}} \\
    & Zero-shot & 55.47 & 50.00 & 56.33 & 81.24 & 64.60 \\
    & Few-shot k=3 & 65.56 & 51.11 & 64.22 & 83.64 & 75.00 \\
    \cmidrule{2-7}
    
    & Zero-shot generation & 68.69 & 49.00 & 53.00 & 86.08 & 65.00 \\
    \cmidrule{2-7}
    
    & \multicolumn{6}{c}{\textbf{Literature-based}} \\
    & \paperonly & 59.22 & 49.00 & 54.00 & 78.80 & 67.68 \\
    & \hyperwrite & 61.63 & 49.67 & 52.67 & 82.36 & 68.76 \\
    & \notebooklm & 53.03 & 49.33 & 51.67 & 68.96 & 62.28 \\
    \cmidrule{2-7}

    & \multicolumn{6}{c}{\textbf{Data-driven}} \\
    & \hypogenic & 75.22 & 81.67 & 68.56 & 82.20 & 76.56 \\
    \cmidrule{2-7}
    & \multicolumn{6}{c}{\textbf{Literature + Data (This work)}}\\
    & \refinemethod  & \bf 77.78 & 55.33 & 63.33 & 89.04 & 78.04 \\
    & Literature $\cup$ \hypogenic & 72.41 & \bf 83.00 & \bf 69.22 & \bf 89.88 & 78.20 \\
    & Literature $\cup$ \refinemethod & 77.19 & 55.33 & 63.00 & 89.52 & \bf 79.24 \\
\midrule

\multirow{13}{*}{\shortstack{\textsc{Llama} \\ 70B-I}} 

    & \multicolumn{6}{c}{\textbf{No hypothesis}}\\
    & Zero-shot & 62.87 & 58.67 & 63.00 & 85.60 & 64.56 \\
    & Few-shot k=3 & 68.56 & 70.45 & 76.00 & 86.80 & 69.44 \\
    \cmidrule{2-7}
    
    & Zero-shot generation & 56.28 & 50.67 & 55.67 & 88.16 & 66.16 \\
    \cmidrule{2-7}
    
    & \multicolumn{6}{c}{\textbf{Literature-based}}\\
    & \paperonly & 64.25 & 50.00 & 49.67 & 80.56 & 66.04 \\
    & \hyperwrite & 58.62 & 50.67 & 54.00 & 83.24 & 74.40 \\
    & \notebooklm & 57.81 & 49.33 & 50.67 & 67.64 & 66.56 \\
    \cmidrule{2-7}
    
    & \multicolumn{6}{c}{\textbf{Data-driven}}\\
    & \hypogenic & 62.06 & 78.67 & 78.00 & 88.44 & 75.48 \\
    \cmidrule{2-7}
    
    & \multicolumn{6}{c}{\textbf{Literature + Data (This work)}}\\
    & \refinemethod & 72.16 & 67.00 & 66.67 & 87.52 & \bf 78.92 \\
    & Literature $\cup$ \hypogenic & \bf 73.72 & \bf 81.33 & \bf 78.67 & 86.72 & 72.56 \\
    & Literature $\cup$ \refinemethod  & 71.75 & 66.67 & 65.67 & \bf 88.76 & 74.80 \\
\bottomrule
\end{tabular}
}
\caption{Accuracy scores on the held-out OOD datasets. Literature + Data outperforms all other methods in every model and task configurations. The bolded numbers outperform the few-shot method ($p<0.05$), as determined by a paired t-test using five random seeds. We show the full results with F1 scores in \cref{tab:ood_table_full}.
}
\label{tab:ood_table_main}
\end{table*}

We first present automatic evaluation results to demonstrate the utility of generated hypotheses for model inference.
We then show that the generated hypotheses are novel and useful, and can improve human decision-making in challenging tasks.

\subsection{Automatic Evaluation}
\label{sec:automatic_results}

\paragraph{Hypotheses generated by combining information from literature and data achieves the best performance across all task and model configurations (\cref{tab:ood_table_main}).} 
First, few-shot inference outperforms zero-shot inference for all task and model configurations, with an average improvement of 6.84\% in accuracy.
In addition, few-shot inference surpasses zero-shot generation and the best of literature-based methods on average accuracy by 7.21\% and 6.78\%, respectively, suggesting off-the-shelve LLMs or literature alone does not generate effective hypotheses for predictive purposes.
In fact, \notebooklm and \hyperwrite
can generate some invalid or irrelevant hypotheses, which degrades their inference performance (see \cref{tab:invalid_examples} in \cref{appendix:hypothesis_examples:examples}).

In contrast, \hypogenic consistently outperforms few-shot inference, improving average accuracy by 5.61\%, highlighting the advantage of data-driven hypotheses. 
Compared to few-shot inference, the hypotheses also offer more interpretable insights.
Furthermore, our best hypothesis generation method combining literature and data outperforms \hypogenic by 3.37\% on average (i.e., an improvement of 11.92\% over few-shot methods and 16.54\% over literature-based methods for \shortgpt, and 6.03\% over few-shot methods and 14.97\% over literature-based methods for \shortllama), demonstrating the benefit of incorporating literature with data.

For \deceptive, \persuasion, and \dreaddit, refining the hypotheses with literature consistently improves inference accuracy compared to \hypogenic, with a 3.92\% improvement on average. On the other hand, refining the hypotheses with literature does not help with \gptgc and \llamagc, but the union of \hypogenic and hypotheses generated from literature consistently performs the best. Comparing with \hypogenic for these two tasks, refining the hypotheses with literature actually results in an accuracy drop by 13.64\%.
This is likely due to that the literature for AIGC detection has relatively few insights on interpretable features to detect AI generated contents, and refining the data-driven hypotheses with that information degrades performance.

\begin{table*}[t]
\centering
\resizebox{0.95\textwidth}{!}{%
\begin{tabular}{@{}p{\textwidth}@{}}
\toprule
\textbf{Case I}: \paperonly and \hypogenic generate \textbf{different} hypotheses\\
\midrule

\textbf{\paperonly:} Deceptive reviews often contain a higher frequency of
first-person singular pronouns, while truthful reviews
may use these pronouns less frequently.\\
\textbf{\hypogenic:} Reviews that reference the reviewer's previous experiences with the hotel brand or similar hotels are more likely to be truthful, while reviews that do not provide any context or comparison to past experiences are more likely to be deceptive. \\

\midrule
\textbf{Case II}: \paperonly and \hypogenic generate \textbf{similar} hypotheses \\
\midrule
\textbf{\paperonly:} Truthful reviews often provide a balanced perspective, while deceptive reviews may seem overly promotional or biased towards a competitor.\\
\textbf{\hypogenic:} Reviews that express a balanced perspective, mentioning both positive and negative aspects of the stay, are more likely to be truthful, whereas reviews that are overly positive or negative without nuance tend to be deceptive. \\
\textbf{\refinemethod:} Reviews that present a balanced perspective by discussing both positive and negative aspects of the stay, particularly with specific examples (e.g., "The location was fantastic, but the air conditioning was broken"), are more likely to be truthful, while reviews that are excessively positive or negative without acknowledging any redeeming qualities (e.g., "This is the best hotel ever!" or "I will never stay here again!") tend to be more deceptive, as they may reflect an attempt to manipulate reader emotions rather than provide an honest assessment. \\
\bottomrule
\end{tabular}
}
\caption{Examples of generated hypotheses from different methods. We show cases where \paperonly and \hypogenic generate different hypotheses or similar hypotheses, and how \refinemethod combines them in the case if they express unifiable ideas.}
\label{tab:hypothesis_with_different_methods}
\end{table*}

To further illustrate our approach, we present a case study of our generated hypotheses in \cref{tab:hypothesis_with_different_methods}.
For most cases, \paperonly and \hypogenic generate different hypotheses as in Case I: one is about first-person singular pronouns, while the other one is about past experiences. 
We include more details on the differences between hypotheses generated by different methods in \cref{sec:human_results}. More examples of hypotheses generated using \unionhyporefine are in \cref{tab:hypotheses-examples}. 
Under some cases, the methods can generate similar hypotheses, and \refinemethod improves the quality of the hypothesis.
In Case II, all three hypotheses focus on balanced perspectives being indicative of truthful reviews.
\refinemethod incorporates the ``reviews that seem to be promoting a competitor'' insight from \paperonly, while also capturing the emphasis on ``lack of nuance'' from \hypogenic. By doing so, \refinemethod offers a more nuanced hypothesis that not only explains how deceptive reviews may manipulate reader emotions, but also provides specific examples to illustrate how balanced perspectives can contribute to truthful assessments. This combination of insights from literature and data allows \refinemethod to offer a more comprehensive and explanatory hypothesis.
We include another case study to compare the generated hypotheses from SciMON \citep{wang2024scimon} and \refinemethod in \cref{appendix:hypothesis_examples:comparing_scimon}, demonstrating the difference between research idea generation and the hypothesis generation task we focus on.
\begin{table*}[t]
\centering
\resizebox{1\textwidth}{!}{%
\begin{tabular}{@{}lllllllll@{}}
\toprule
Generation Model & Inference Model & \deceptive & \llamagc & \gptgc & \persuasion & \dreaddit \\
\cmidrule(lr){3-3} \cmidrule(lr){4-4} \cmidrule(lr){5-5} \cmidrule(lr){6-6} \cmidrule(lr){7-7}
& & OOD Accuracy & OOD Accuracy & OOD Accuracy & OOD Accuracy & OOD Accuracy \\
\midrule

\shortgpt & \shortgpt & 77.78 & 83.00 & 69.22 & 89.88 & 79.24 \\
& \shortllama & 72.53 (\decrease $5.25$) & 71.67 (\decrease $11.33$) & 76.33 (\increase $7.11$)& 86.88 (\decrease $3.00$) & 72.36 (\decrease $6.88$) \\
\shortllama & \shortllama & 73.72 & 81.33 & 78.67 & 88.76 & 78.92 \\
& \shortgpt & 70.31 (\decrease $3.41$) & 57.00 (\decrease $24.33$) & 74.67 (\decrease $4.00$) & 89.36 (\increase $0.60$) & 77.28 (\decrease $1.64$) \\
\midrule

Generation Model & Inference Model & IND Accuracy & IND Accuracy & IND Accuracy & IND Accuracy & IND Accuracy \\
\cmidrule(lr){3-3} \cmidrule(lr){4-4} \cmidrule(lr){5-5} \cmidrule(lr){6-6} \cmidrule(lr){7-7}
\shortgpt & \shortgpt & 70.76 & 73.00 & 68.33 & 90.52 & 70.88 \\
& \shortllama & 62.20 (\decrease $8.56$) & 69.00 (\decrease $4.00$) & 81.67 (\increase $13.34$) & 90.40 (\decrease $0.12$) & 74.88 (\increase $4.00$) \\
\shortllama & \shortllama & 72.60 & 78.33 & 80.67 & 91.24 & 78.68 \\
& \shortgpt & 66.28 (\decrease $6.32$) & 68.00 (\decrease $10.33$) & 68.33 (\decrease $12.34$) & 89.88 (\decrease $1.36$) & 68.60 (\decrease $10.08$) \\

\bottomrule
\end{tabular}
}
\caption{Cross-model inference performance.}
\label{tab:cross_model_all}
\end{table*}

\paragraph{Performance on IND held-out datasets.} 
Similar to \cref{tab:ood_table_main}, combining literature and data achieves the best accuracy and F1 scores in most cases on the held-out IND datasets (see \cref{tab:ind_table_main} in the Appendix). For some cases, such as using Llama on the IND datasets for \gptgc, \llamagc, and \persuasion, \hypogenic gets the top performance compared to other methods. This is not surprising, since \hypogenic generates hypotheses by looking at IND data only. In contrast, our methods that take information from both literature and data may generate hypotheses that 
are more generally applicable but with slightly worse performance on the IND data. 
Thus in \cref{tab:ood_table_main}, the hypotheses generated from both literature and data performs the best on all methods for OOD datasets.

\paragraph{Generated hypotheses can be effectively transferred to a different model.} To further check the generalization behavior of our generated hypotheses, we take the hypotheses from the best-performing method with our literature+data approach and then use the other model to perform inference.
\cref{tab:cross_model_all} shows that the generated hypotheses from one model remain effective for the other model, the performance exhibits no significant change in most cases (drop $<$5\% in 11 out of 20 cases). 
Even with this performance drop, our methods still outperform the few-shot baseline by 3.76\% and 3.66\% in OOD and IND settings.
This finding further demonstrates the robustness of our hypothesis generation and 
inference.

A significant outlier case is for \llamagc OOD: when using \shortllama-generated hypotheses for \gptgc OOD and ask \shortgpt to perform hypothesis-based inference, the inference performance can degrade significantly. This can be due to innate deficits in the task setting, as LLMs tend to favor and better detect their own writing \citep{panickssery2024llmevaluatorsrecognizefavor}. 

\paragraph{Our hypothesis generation method is robust to different prompts and hyperparameters, and it is effective for smaller models.} We conduct additional experiments to test the robustness of our method with different prompts, hyperparameters, and with \textsc{Llama-3.1-8B-Instruct}. For the OOD setting, in \cref{tab:prompt_sensitivity_ood} and \cref{tab:hyperparameter_search_ood}, we show that the average performance of our method only drop by 0.20\% and 0.07\% with different prompts and hyperparameters, respectively. In \cref{tab:llama_8b_results}, we further show that with \textsc{Llama-3.1-8B-Instruct}, our literature + data approach outperforms all baselines, with an average accuracy improvement of 15.27\%, 13.04\%, and 4.88\% over few-shot, literature-based methods, and \hypogenic, respectively. These results highlight the robustness of our method and its scalability and reproducibility with smaller models. Full details are in \cref{appendix:additional_experiments}.

\subsection{Human Evaluation}
\label{sec:human_results}

\paragraph{Generated hypotheses improve human decision-making in both AIGC Detection and Deception Detection.}
We experiment with the \gptgc task for AIGC Detection, and the average human accuracy improves by 14.19\% (58.86\% $\rightarrow$ 73.05\%) when we provide hypotheses as assistance. 
We perform a statistical t-test and obtain a p-value of 0.01, indicating that the improvement is significant.
In Deception Detection, the introduction of hypotheses boosts human accuracy by 7.44\% (57.14\% $\rightarrow$ 64.58\%), with a p-value of 0.04. 

\begin{table}[t]
\centering
\resizebox{0.5\textwidth}{!}{%
\small
\begin{tabular}{lccc}
\toprule
 & Clarity & Novelty & Plausibility \\

\midrule
\refinemethod & $4.22\pm0.07$ & $2.85\pm0.10$ & $4.21\pm0.08$ \\
\midrule
\notebooklm & $3.71\pm0.09*$ & $3.01\pm0.10$ & $3.65\pm0.10*$ \\

\hyperwrite & $3.08\pm0.12*$ & $2.15\pm0.11*$ & $3.32\pm0.13*$ \\
\bottomrule
\end{tabular}
}
\caption{Human ratings on hypotheses generated by \refinemethod, \notebooklm, and \hyperwrite. * indicates that the difference between the rating with \refinemethod is statistically significant ($p<$0.001).
}
\label{tab:likert-results}
\end{table}

\begin{table*}[t]
\centering
\resizebox{0.95\textwidth}{!}{%
\begin{tabular}{p{12cm}c}
\toprule
\textbf{Hypotheses} & \textbf{Frequency of Selection} \\
\midrule
\textbf{Hypothesis 1:} AI-generated texts tend to use more elaborate and descriptive language, including adjectives and adverbs, to create a sense of atmosphere and immersion. Human-written texts, on the other hand, tend to be more concise and straightforward in their language use. & 38.79\% \\
\textbf{Hypothesis 2:} Human-written texts are more likely to contain errors or idiosyncrasies in grammar and punctuation, reflecting the natural imperfections of human writing, while AI-generated texts typically maintain a higher level of grammatical accuracy. & 34.55\% \\
\textbf{Hypothesis 3:} Human-written texts tend to have a more conversational tone and colloquial language, while AI-generated texts tend to be more formal and lack idiomatic expressions. & 44.55\% \\
\textbf{No hypothesis selected} & 3.94\% \\
\bottomrule
\end{tabular}
}
\caption{How often participants use hypotheses in AIGC Detection. We allow users to select multiple hypotheses for each instance they make prediction on, so the total frequency can exceed 100\%.}
\label{tab:utility-aigc-detection}
\end{table*}

When hypotheses are present, participants would use them to assist decision-making for over 90\% of the time.
All three presented hypotheses are selected to be used with frequency greater than 30\% (\cref{tab:utility-aigc-detection}, \cref{tab:utility-deception-detection} in the Appendix).
For example, the most used hypothesis, with frequency of 44.55\%, in AIGC detection is ``Human-written texts tend to have a more conversational tone and colloquial language, while AI-generated texts tend to be more formal and lack idiomatic expressions.''
For both tasks, 100\% of the participants find the hypotheses to be helpful, and over 40\% find them to be ``Very helpful'' or ``Extremely helpful''.

\paragraph{\refinemethod hypotheses are rated higher in clarity and plausibility compared to those generated by existing methods (\cref{tab:likert-results}).}
Hypotheses generated by \refinemethod achieve statistically significantly higher clarity and plausibility scores than those generated by \notebooklm and \hyperwrite. In terms of novelty, \notebooklm receives slightly higher ratings; however, the difference between \notebooklm and \refinemethod is not statistically significant. 
Recall that hypotheses generated by \notebooklm do not have strong predictive power.
In other words, 
generating ``novel'' hypotheses is easier if they are not constrained by plausibility.

\paragraph{Humans rate literature-based and data-driven hypotheses as distinct.}
We assign novelty labels to hypothesis pairs based on a majority vote from three human annotators, who evaluate whether a hypothesis give meaningfully different information (``novel'') from another. 84\% and 80\% of the hypotheses are rated novel for Deception Detection and AIGC Detection, respectively, demonstrating the complementarity between literature-based and data-driven approaches.

\section{Related Work}
\label{sec:related_work}

\paragraph{Literature-based research idea generation.}
\citet{baek2024researchagent}, \citet{wang2024scimon}, and \citet{ghafarollahi2024sciagentsautomatingscientificdiscovery} use LLMs to build knowledge graphs from literature and generate research ideas, such as proposing new problem setups, methodologies, or evaluation frameworks. Unlike their focus on ideation, our work generates hypotheses to explain a phenomenon with real observations (see \cref{appendix:hypothesis_examples:comparing_scimon} for detailed comparison). \citet{yang2024moose} generates hypotheses from raw web data but relies on annotated hypotheses from literature. These methods 
require extensive adaptation, so we developed our own literature-based approach.

\paragraph{Data-driven hypothesis generation.}
Besides \hypogenic, we review additional works 
on discovering unseen patterns from data.
\citet{zhong2023goal} discovers patterns by analyzing difference between large corpora. 
\citet{pham2023topicgpt} makes discovery by generating and refining interpretable topics. \citet{funsearch} uncovers new solutions in open math problems by iteratively updating programs. \citet{qiu2024phenomenal} and \citet{yanginductivereasoners} evaluate LLMs' ability in performing inductive reasoning in synthetic settings. 
\citet{batista2024words} uses LLMs to generate hypotheses and conducts comprehensive experiments to study human engagements with headlines. 
We choose \hypogenic as the backbone for data-driven hypothesis generation as their tasks are most similar to ours, and their approach to hypothesis updates integrates naturally into our refinement process.

\paragraph{Automated scientific research with LLMs.}
There is growing interest in developing LLM-powered methods and multi-agent frameworks to assist scientific research. 
\citet{aiscientist} designs an LLM agent to generate full research papers.
\citet{li2024mlrcopilotautonomousmachinelearning} proposes a method to generate research ideas from existing literature and automatically implement and execute experiments.
In contrast, our work focuses primarily on hypothesis generation, as we believe it is crucial to preserve human agency and oversight in the scientific research process.

To evaluate LLM generated hypotheses, \citet{qi2023large} examines whether they contain novel information not found in existing literature.
\citet{si2024llmsgeneratenovelresearch} asks experts to rate the novelty of LLM-proposed research ideas in the NLP domain.
While these studies highlight LLMs' ability to generate novel hypotheses, they do not conduct human subject experiments to validate the effectiveness of hypotheses.
To this end, we conduct the first human study to test the utility of LLM-generated hypotheses in supporting human decision-making.

Significant efforts have also been made to evaluate and benchmark 
multi-agent frameworks on data analysis tasks \citep{majumder2024discoverybenchdatadrivendiscoverylarge, gu2024bladebenchmarkinglanguagemodel, InfiAgent-DABench, chen2024scienceagentbenchrigorousassessmentlanguage, MLAgentBench, guo2024dsagent}, 
literature processing and information retrieval tasks \citep{press2024citemelanguagemodelsaccurately, ajith2024litsearchretrievalbenchmarkscientific, kang2024researcharenabenchmarkingllmsability, zhang2024masswnewdatasetbenchmark}, and more general research tasks \citep{tian2024scicoderesearchcodingbenchmark,jansen2024discoveryworldvirtualenvironmentdeveloping}.

\section{Conclusion}
\label{sec:conclusion}
We propose a novel approach that integrates literature and data to generate hypotheses, with extensive and systematic evaluations. Our method consistently outperforms all baselines, including existing literature-based and data-driven approaches. Furthermore, human evaluations reveal that our generated hypotheses also improve human decision-making in challenging tasks.

\section{Limitations}
\label{sec:limitations}

Our automated evaluation uses two recent models on datasets across various domains, showing the effectiveness of our method across diverse settings. However, we did not further evaluate our hypotheses on some tasks that require representations beyond natural language, such as math problem solving and code generation.

The literature corpus used for literature-based hypothesis generation is limited in terms of size and collection method. The collection is carried out by manually searching and collecting up to 10 papers on Semantic Scholar or Google Scholar.
Though with the limited literature corpus we already show that our methods yield competent performance, a natural future direction is to enhance the literature component with automatic and scalable retrieval.

Similarly, we achieved satisfactory performance across different models and tasks with the initial set of hyperparameters. However, we did not perform an exhaustive hyperparameter search, which may have yielded further enhancements to the performance of our methods. This represents a limitation of our study that could be addressed in future work.

Our experiments with human subjects is a proof of concept.
The number of participants in our human evaluation is relatively small.
As a result, we do not believe that we have the statistical power to distinguish, for example, the difference between \hypogenic and \refinemethod.
Although this is not the focus of our study,
we encourage future work to conduct large-scale experiments in focused domains to validate the hypotheses generated through human-AI collaboration.

Last but not least, we manually chose three hypotheses through ablation-style study and subjective judgment for experiments with human subjects.
We believe this process is the essence of human-AI collaboration in future scientific processes.
It requires future exploration to identify the optimal collaboration regime.

\section*{Acknowledgments}
We thank members of the Chicago Human+AI Lab for their helpful comments.
We also thank the anonymous participants on Prolific for participating in our study.
This work is supported in part by NSF grants IIS-2126602 and an award from the Sloan Foundation.

\bibliography{custom.bib}

\clearpage
\appendix

\section{Prompts}
\label{appendix:prompts}

All our prompts for LLMs are separated into system prompts and user prompts.  System prompts contain role and tone information, followed by detailed descriptions of the task and the expected response format. User prompts contain useful information for hypothesis generation, refinement, or inference, including information from literature, instances from datasets, and previously generated hypotheses. Below are some examples of the prompts that we use for each task. 

\subsection{Deception Detection}
\label{appendix:prompts:deceptive_reviews}
\begin{lstlisting}[caption={Data-Based Hypothesis Generation with HypoGeniC.},label={lst:hypogenic:review},firstnumber=auto,xleftmargin=0pt]
@\textcolor{mygray}{System Prompt}@
You're a professional hotel review analyst.
Given a set of hotel reviews, we want to generate hypotheses that are useful for predicting whether a review is truthful or deceptive. In other words, we want to know whether the review is written by a someone who actually lived in the hotel.

Using the given examples, please propose @\textcolor{mypurple}{<num\_hypotheses>}@ possible hypothesis pairs.
These hypotheses should identify specific patterns that occur across the provided reviews.

Each hypothesis should contain a pair of the following:
a. A hypothesis about what makes reviews more likely to be truthful
b. The opposite hypothesis about what makes reviews more likely to be deceptive 

Generate them in the format of 1. [hypothesis], 2. [hypothesis], ... @\textcolor{mypurple}{<num\_hypotheses>}@. [hypothesis].
The hypotheses should analyze what kind of reviews are likely to be truthful or deceptive.

@\textcolor{mygray}{User Prompt}@
We have seen some hotel reviews:
@\textcolor{mygray}{··· more examples here ···}@
Please generate hypotheses that are useful for predicting whether a review is truthful or deceptive. 
Propose @\textcolor{mypurple}{<num\_hypotheses>}@ possible hypotheses. Generate them in the format of 1. [hypothesis], 2. [hypothesis], ... @\textcolor{mypurple}{<num\_hypotheses>}@. [hypothesis].
Proposed hypotheses:
\end{lstlisting}

\begin{lstlisting}[caption={Literature-Based Hypothesis Generation.},label={lst:litgen:review},firstnumber=auto,xleftmargin=0pt]
@\textcolor{mygray}{System Prompt}@
You're a professional hotel review analyst.
Given some key findings from a series of research papers, we want to generate hypotheses that are useful for predicting whether a review is truthful or deceptive. In other words, we want to know whether the review is written by a someone who actually lived in the hotel.

Using the given relevant literatures, please propose @\textcolor{mypurple}{<num\_hypotheses>}@ possible hypothesis pairs.
These hypotheses should identify specific patterns that occur across the provided reviews.

Each hypothesis should contain a pair of the following:
a. A hypothesis about what makes reviews more likely to be truthful
b. The opposite hypothesis about what makes reviews more likely to be deceptive 

Generate them in the format of 1. [hypothesis], 2. [hypothesis], ... @\textcolor{mypurple}{<num\_hypotheses>}@. [hypothesis].
The hypotheses should analyze what kind of reviews are likely to be truthful or deceptive.

@\textcolor{mygray}{User Prompt}@
We have some key findings from a series of research papers that might be useful for generating the required @\textcolor{mypurple}{<num\_hypotheses>}@. hypotheses:
@\textcolor{mygray}{··· information from literature here ···}@
Please generate hypotheses that are useful for predicting whether a review is truthful or deceptive. 
When generating hypotheses, remember not to overuse your own knowledge. Always refer to the key findings from research papers provided. Directly cite passages in the key findings when generating a hypothesis.
Propose @\textcolor{mypurple}{<num\_hypotheses>}@ possible hypotheses. Remember to generate @\textcolor{mypurple}{<num\_hypotheses>}@ hypotheses! Generate them in the format of 1. [hypothesis], 2. [hypothesis], ... @\textcolor{mypurple}{<num\_hypotheses>}@. [hypothesis].
Proposed hypotheses:
\end{lstlisting}

\begin{lstlisting}[caption={Paper Summarization.},label={lst:summarize:review},firstnumber=auto,xleftmargin=0pt]
@\textcolor{mygray}{System Prompt}@
You are a helpful assistant for summarizing key findings in research papers on a given topic.

@\textcolor{mygray}{User Prompt}@
Summarize the following research paper, focusing ONLY on this question: What is useful for one to decide whether a review is truthful or deceptive in real life? 
Focus on hypotheses of what kind of reviews tend to be deceptive, do not include technical details in the paper. 
@\textcolor{mygray}{... literature texts here ...}@
\end{lstlisting}

\begin{lstlisting}[caption={Hypothesis Refinement Based on Data.},label={lst:data_refine:review},firstnumber=auto,xleftmargin=0pt]
@\textcolor{mygray}{System Prompt}@
You're a social scientist working on a project to identify deceptive hotel reviews.
Given a set of hotel reviews, we want to generate hypotheses that are useful for predicting whether a review is truthful or deceptive. In other words, we want to know whether the review is written by a someone who actually lived in the hotel.

Using the given examples, refine the hypothesis pairs provided.
The desired hypotheses should identify specific patterns that occur across the provided reviews.

Each hypothesis should contain a pair of the following:
a. A hypothesis about what makes reviews more likely to be truthful
b. The opposite hypothesis about what makes reviews more likely to be deceptive

Generate refined hypotheses in the format of 1. [hypothesis], 2. [hypothesis], ... @\textcolor{mypurple}{<num\_hypotheses>}@. [hypothesis].
The hypotheses should analyze what kind of reviews are likely to be truthful or deceptive.

@\textcolor{mygray}{User Prompt}@
We have seen some hotel reviews:
@\textcolor{mygray}{··· more examples here ···}@
We have some hypotheses need to be refined:
@\textcolor{mygray}{... hypotheses to be refined here ...}@
Please refine these hypotheses to make them more specific and useful for predicting whether a review is truthful or deceptive. 
When refining the hypotheses, feel free to change the key information or topic of a hypothesis based on the provided prevailing patterns in data if you think it is necessary.
Generate refined hypotheses in the format of 1. [hypothesis], 2. [hypothesis], ... @\textcolor{mypurple}{<num\_hypotheses>}@. [hypothesis].
Refined hypotheses:

\end{lstlisting}

\begin{lstlisting}[caption={Hypothesis Refinement Based on Literature.},label={lst:lit_refine:review},firstnumber=auto,xleftmargin=0pt]
@\textcolor{mygray}{System Prompt}@
You're a social scientist working on a project to identify deceptive hotel reviews.
Given a set of hotel reviews, we want to generate hypotheses that are useful for predicting whether a review is truthful or deceptive. In other words, we want to know whether the review is written by a someone who actually lived in the hotel.

Using the given relevant literatures, refine the hypothesis pairs provided.
The desired hypotheses should identify specific patterns that occur across the provided reviews.

Each hypothesis should contain a pair of the following:
a. A hypothesis about what makes reviews more likely to be truthful
b. The opposite hypothesis about what makes reviews more likely to be deceptive 

Generate refined hypotheses in the format of 1. [hypothesis], 2. [hypothesis], ... @\textcolor{mypurple}{<num\_hypotheses>}@. [hypothesis].
The hypotheses should analyze what kind of reviews are likely to be truthful or deceptive.

@\textcolor{mygray}{User Prompt}@
We have some key findings from a series of research papers that might be useful for generating hypotheses:
@\textcolor{mygray}{··· information from literature here ···}@
We have some hypotheses need to be refined:
@\textcolor{mygray}{... hypotheses to be refined here ...}@
Please refine these hypotheses to make them more specific and useful for predicting whether a review is truthful or deceptive. 
When refining the hypotheses, feel free to change the key information or topic of a hypothesis based on the provided key findings if you think it is necessary.
Generate refined hypotheses in the format of 1. [hypothesis], 2. [hypothesis], ... @\textcolor{mypurple}{<num\_hypotheses>}@. [hypothesis].
Refined hypotheses:
\end{lstlisting}

\begin{lstlisting}[caption={Multiple-Hypothesis-Based Inference.},label={lst:inference:review},firstnumber=auto,xleftmargin=0pt]
@\textcolor{mygray}{System Prompt}@
You are a professional deceptive detection agent and your job is to determine whether a hotel review is truthful or deceptive. 
In other words, we want to know whether the review is written by someone who had real experiences with the hotel. 
From past experiences, you learned some patterns. 
You need to determine whether each of the patterns holds for the current hotel review, and also predict whether the current hotel review is truthful or deceptive. 
Give an answer. The answer should be one word (truthful or deceptive).
Give your final answer in the format of {Final answer: answer}

@\textcolor{mygray}{User Prompt}@
Our learned patterns: @\textcolor{mypurple}{<generated\_hypotheses>}@     
A hotel review is the following: @\textcolor{mypurple}{<review>}@

Given the pattern you learned above, give an answer of whether the hotel review above is deceptive or truthful.
Think step by step.
First step: Think about which pattern can be applied to the hotel review.
Second step: Based on the pattern, is this hotel review deceptive or truthful?
Final step: give your final answer in the format of {Final answer: answer}
\end{lstlisting}

\subsection{AI-Generated Content (AIGC) Detection}
\label{appendix:prompts:aigc}

\begin{lstlisting}[caption={AIGC Detection Dataset Generation.}, label={lst:gptgc_generation:aigc},firstnumber=auto,xleftmargin=0pt]

@\textcolor{mygray}{System Prompt}@
You are a professional writer. 
You will be given a story prompt. Your task is to write a coherent and creative story based on the prompt. Ensure the story has a clear structure, engaging characters, and stays true to the prompt's core idea. Keep the language clear and appropriate to the story's tone.

@\textcolor{mygray}{User Prompt}@
@\textcolor{mygray}{... story-writing prompt here ...}@
@\textcolor{mygray}{example: }@
[ WP ] You 've been able to read minds since you turned 7 . Mostly you watch people 's thoughts passively and undetected but one day someone talks back .\n
\end{lstlisting}

\begin{lstlisting}[caption={Data-Based Hypothesis Generation with HypoGeniC.},label={lst:hypogenic:aigc},firstnumber=auto,xleftmargin=0pt]
@\textcolor{mygray}{System Prompt}@
You're a an AI generated content detection expert. You are great at detecting what type of text is generated by AI.
Given a set of texts, we want to generate hypotheses that are useful for predicting whether a piece of text is generated by AI. In other words, we want to know whether the text is written by a human or generated by AI.

Your task is to identify what patterns or traits show up more in AI generated texts, and what shows up more in human written texts.  Focus on the generalizable insight that can be applied in other contexts. Ignore things that are specific to this story. Do not make references this story they may not be for others.

Using the given examples, please propose @\textcolor{mypurple}{<num\_hypotheses>}@ possible hypothesis pairs.
When proposing hypothesis, look closely into the given examples and identify specific patterns that occur across the provided text examples. 
The hypotheses should be clear, easy to understand, and have specific details such that one can apply the hypotheses to predict whether a piece of text is written by human or AI.

Generate them in the format of 1. [hypothesis], 2. [hypothesis], ... @\textcolor{mypurple}{<num\_hypotheses>}@. [hypothesis].
The hypotheses should analyze what kind of text is likely to be written by human or AI.

@\textcolor{mygray}{User Prompt}@
We have seen some texts:
@\textcolor{mygray}{... more examples here ...}@
Please generate hypotheses that are useful for predicting predicting whether a piece of text is written by human or AI.
Propose @\textcolor{mypurple}{<num\_hypotheses>}@ possible hypotheses. Generate them in the format of 1. [hypothesis], 2. [hypothesis], ... @\textcolor{mypurple}{<num\_hypotheses>}@. [hypothesis].

When proposing hypothesis, look closely into the given examples and identify specific patterns that occur across the provided text examples. 

Please make sure that the hypotheses are:
i. clear (i.e., precise , not too wordy , and easy to understand);
ii. generalizable to novel situations (i.e., they would make sense if applied to other AI generated content detection experiments or other messaging contexts);
iii. empirically plausible (i.e., this is a dimension on which messages can vary on);
iv. unidimensional (i.e., avoid hypotheses that list multiple constructs so if there are many things changing , pick one);
v. usable (i.e., a human equipped with this insight could use it to predict if a new piece of text is generated AI in a similar way)

Proposed hypotheses:
\end{lstlisting}

\begin{lstlisting}[caption={Literature-Based Hypothesis Generation.},label={lst:litgen:aigc},firstnumber=auto,xleftmargin=0pt]
@\textcolor{mygray}{System Prompt}@
You're a professional AI content detector.
Given some key findings from a series of research papers, we want to generate hypotheses that are useful for detecting whether a piece of text is written by human or AI. 

Your task is to identify what patterns or traits show up more in AI generated texts, and what shows up more in human written texts.  Focus on the generalizable insight that can be applied in other contexts. Ignore things that are specific to this story. Do not make references this story they may not be for others.

Using the given relevant literatures, please propose @\textcolor{mypurple}{<num\_hypotheses>}@ possible hypothesis pairs.
These hypotheses should identify specific patterns that occur across the provided texts.

Generate them in the format of 1. [hypothesis], 2. [hypothesis], ... @\textcolor{mypurple}{<num\_hypotheses>}@. [hypothesis].
The hypotheses should analyze what kind of text is likely to be written by human or AI.

@\textcolor{mygray}{User Prompt}@
We have some key findings from a series of research papers that might be useful for generating the required @\textcolor{mypurple}{<num\_hypotheses>}@ hypotheses:
@\textcolor{mygray}{··· information from literature here ···}@
Please generate hypotheses that are useful for predicting whether a piece of text is written of human or AI. 
Propose @\textcolor{mypurple}{<num\_hypotheses>}@ possible hypotheses. Remember to generate @\textcolor{mypurple}{<num\_hypotheses>}@ hypotheses! Generate them in the format of 1. [hypothesis], 2. [hypothesis], ... @\textcolor{mypurple}{<num\_hypotheses>}@. [hypothesis].
Proposed hypotheses:
\end{lstlisting}

\begin{lstlisting}[caption={Paper Summarization.},label={lst:summarize:aigc},firstnumber=auto,xleftmargin=0pt]
@\textcolor{mygray}{System Prompt}@
You are a helpful assistant for summarizing key findings in research papers on a given topic.

@\textcolor{mygray}{User Prompt}@
Summarize the following research paper, focusing ONLY on this question: What is useful for one to detect whether some text is generated by AI? 
Focus on hypotheses of what kind of text tend to be generated by AI, do not include technical details in the paper. 
@\textcolor{mygray}{... literature texts here ...}@
\end{lstlisting}

\begin{lstlisting}[caption={Hypothesis Refinement Based on Data.},label={lst:data_refine:aigc},firstnumber=auto,xleftmargin=0pt]
@\textcolor{mygray}{System Prompt}@
You're a an AI generated content detection expert. You are great at detecting what type of text is generated by AI.
Given a set of texts, we want to generate hypotheses that are useful for predicting whether a piece of text is generated by AI. In other words, we want to know whether the text is written by a human or generated by AI.

Using the given examples, refine the hypothesis pairs provided.
The desired hypotheses should identify specific patterns that occur across the provided text examples.

Generate refined hypotheses in the format of 1. [hypothesis], 2. [hypothesis], ... @\textcolor{mypurple}{<num\_hypotheses>}@. [hypothesis].
The hypotheses should analyze what kind of text is likely to be written by human or AI.

@\textcolor{mygray}{User Prompt}@
We have seen some texts:
@\textcolor{mygray}{··· more examples here ···}@
We have some hypotheses need to be refined:
@\textcolor{mygray}{... hypotheses to be refined here ...}@
Please refine these hypotheses to make them more specific and useful for predicting whether a piece of text is written by human or AI.
When refining the hypotheses, feel free to change the key information or topic of a hypothesis based on the provided prevailing patterns in data if you think it is necessary.
Generate refined hypotheses in the format of 1. [hypothesis], 2. [hypothesis], ... @\textcolor{mypurple}{<num\_hypotheses>}@. [hypothesis].
Refined hypotheses:
\end{lstlisting}

\begin{lstlisting}[caption={Hypothesis Refinement Based on Literature.},label={lst:lit_refine:aigc},firstnumber=auto,xleftmargin=0pt]
@\textcolor{mygray}{System Prompt}@
You're a an AI generated content detection expert. You are great at detecting what type of text is generated by AI.
Given a set of texts, we want to generate hypotheses that are useful for predicting whether a piece of text is generated by AI. In other words, we want to know whether the text is written by a human or generated by AI.

Using the given relevant literatures, refine the hypothesis pairs provided.
The desired hypotheses should identify specific patterns that occur across the provided text examples.

Generate refined hypotheses in the format of 1. [hypothesis], 2. [hypothesis], ... @\textcolor{mypurple}{<num\_hypotheses>}@. [hypothesis].
The hypotheses should analyze what kind of text is likely to be written by human or AI.

@\textcolor{mygray}{User Prompt}@
We have some key findings from a series of research papers that might be useful for generating hypotheses:
@\textcolor{mygray}{··· information from literature here ···}@
We have some hypotheses need to be refined:
@\textcolor{mygray}{... hypotheses to be refined here ...}@
Please refine these hypotheses to make them more specific and useful for predicting whether a piece of text is written by human or AI.
When refining the hypotheses, feel free to change the key information or topic of a hypothesis based on the provided prevailing patterns in data if you think it is necessary.
Generate refined hypotheses in the format of 1. [hypothesis], 2. [hypothesis], ... @\textcolor{mypurple}{<num\_hypotheses>}@. [hypothesis].
Refined hypotheses:
\end{lstlisting}

\begin{lstlisting}[caption={Multiple-Hypothesis-Based Inference.},label={lst:inference:aigc},firstnumber=auto,xleftmargin=0pt]
@\textcolor{mygray}{System Prompt}@
You are an AI generated content detection agent and want to determine whether a piece of text is written by a human or generated by an AI. In other words, we want to know whether the text is written by a human or generated by AI.
From past experiences, you learned some patterns. 
You need to determine whether each of the patterns holds for the current text, and also predict whether the current text is written by human or AI. 
Give an answer. The answer should be one word (AI or HUMAN).
Give your final answer in the format of "Final answer: ANSWER"

@\textcolor{mygray}{User Prompt}@
Our learned patterns: @\textcolor{mypurple}{<generated\_hypotheses>}@
New text:
Here is a story: @\textcolor{mypurple}{<story>}@

Given the patterns you learned above, give an answer of whether the current text is written by human or AI. 
Think step by step.
First step: Think about which pattern can be applied to the story.
Second step: Based on the pattern, is this story written by human or AI?
You must give your final answer in the format of "Final answer: ANSWER".
\end{lstlisting}

\subsection{Mental Stress Detection}
\label{appendix:prompts:dreaddit}

\begin{lstlisting}[caption={Data-Based Hypothesis Generation with HypoGeniC.},label={lst:hypogenic:dreaddit},firstnumber=auto,xleftmargin=0pt]
@\textcolor{mygray}{System Prompt}@
You're a psychologist and social scientist studying people's stress and their online posts.
given a set of reddit posts, we want to generate hypotheses that are useful for deciding people's stress status (has stress or no stress) based on reddit post.

Using the given examples, please propose @\textcolor{mypurple}{<num\_hypotheses>}@ possible hypothesis pairs.
These hypotheses should identify specific patterns that occur across the provided posts.

Each hypothesis should contain a pair of the following:
a. A hypothesis about what makes the post more likely to indicate that the poster has stress
b. The opposite hypothesis about what makes the post more likely to indicate that the poster does not have stress

Generate them in the format of 1. [hypothesis], 2. [hypothesis], ... @\textcolor{mypurple}{<num\_hypotheses>}@. [hypothesis].
The hypotheses should analyze what kind of posts are likely to indicate stress or no stress.

@\textcolor{mygray}{User Prompt}@
We have seen some reddit posts:
@\textcolor{mygray}{··· more examples here ···}@
Please generate hypotheses that are useful for deciding people's stress status (has stress or no stress) based on reddit post.
Propose @\textcolor{mypurple}{<num\_hypotheses>}@ possible hypotheses. Generate them in the format of 1. [hypothesis], 2. [hypothesis], ... @\textcolor{mypurple}{<num\_hypotheses>}@. [hypothesis].
Proposed hypotheses:
\end{lstlisting}

\begin{lstlisting}[caption={Literature-Based Hypothesis Generation.},label={lst:litgen:dreaddit},firstnumber=auto,xleftmargin=0pt]
@\textcolor{mygray}{System Prompt}@
You're a psychologist and social scientist studying people's stress and their online posts.
Given some key findings from a series of research papers, we want to generate hypotheses that are useful for deciding people's stress status (has stress or no stress) based on reddit post.

Using the given relevant literatures, please propose @\textcolor{mypurple}{<num\_hypotheses>}@ possible hypothesis pairs.
These hypotheses should identify specific patterns that occur across the provided posts.

Each hypothesis should contain a pair of the following:
a. A hypothesis about what makes the post more likely to indicate that the poster has stress
b. The opposite hypothesis about what makes the post more likely to indicate that the poster does not have stress

Generate them in the format of 1. [hypothesis], 2. [hypothesis], ... @\textcolor{mypurple}{<num\_hypotheses>}@. [hypothesis].
The hypotheses should analyze what kind of posts are likely to indicate stress or no stress.

@\textcolor{mygray}{User Prompt}@
We have some key findings from a series of research papers that might be useful for generating the required @\textcolor{mypurple}{<num\_hypotheses>}@ hypotheses:
@\textcolor{mygray}{··· information from literature here ···}@
Please generate hypotheses that are useful for deciding people's stress status (has stress or no stress) based on reddit post.
Propose @\textcolor{mypurple}{<num\_hypotheses>}@ possible hypotheses. Remember to generate @\textcolor{mypurple}{<num\_hypotheses>}@ hypotheses! Generate them in the format of 1. [hypothesis], 2. [hypothesis], ... @\textcolor{mypurple}{<num\_hypotheses>}@. [hypothesis].
Proposed hypotheses:
\end{lstlisting}

\begin{lstlisting}[caption={Paper Summarization.},label={lst:summarize:dreaddit},firstnumber=auto,xleftmargin=0pt]
@\textcolor{mygray}{System Prompt}@
You are a helpful assistant for summarizing key findings in research papers on a given topic.

@\textcolor{mygray}{User Prompt}@
Summarize the following research paper, focusing ONLY on this question: What is useful for one to judge whether a reddit poster has stress based on one of their reddit post content?
Focus on hypotheses of what kind of posts indicate stress, do not include technical details in the paper. 
@\textcolor{mygray}{... literature texts here ...}@
\end{lstlisting}

\begin{lstlisting}[caption={Hypothesis Refinement Based on Data.},label={lst:data_refine:dreaddit},firstnumber=auto,xleftmargin=0pt]
@\textcolor{mygray}{System Prompt}@
You're a psychologist and social scientist working on a project to identify whether a person has stress based on reddit posts.
given a set of reddit posts, we want to generate hypotheses that are useful for deciding people's stress status (has stress or no stress) based on reddit post.

Using the given examples, refine the hypothesis pairs provided.
The desired hypotheses should identify specific patterns that occur across the provided posts.

Each hypothesis should contain a pair of the following:
a. A hypothesis about what makes the post more likely to indicate that the poster has stress
b. The opposite hypothesis about what makes the post more likely to indicate that the poster does not have stress

Generate refined hypotheses in the format of 1. [hypothesis], 2. [hypothesis], ... @\textcolor{mypurple}{<num\_hypotheses>}@. [hypothesis].
The hypotheses should analyze what kind of posts are likely to indicate stress or no stress.

@\textcolor{mygray}{User Prompt}@
We have seen some reddit posts:
@\textcolor{mygray}{··· more examples here ···}@
We have some hypotheses need to be refined:
@\textcolor{mygray}{... hypotheses to be refined here ...}@
Please refine these hypotheses to make them more specific and useful for deciding people's stress status (has stress or no stress) based on reddit post.
Generate refined hypotheses in the format of 1. [hypothesis], 2. [hypothesis], ... @\textcolor{mypurple}{<num\_hypotheses>}@. [hypothesis].
Refined hypotheses:
\end{lstlisting}

\begin{lstlisting}[caption={Hypothesis Refinement Based on Literature.},label={lst:lit_refine:dreaddit},firstnumber=auto,xleftmargin=0pt]
@\textcolor{mygray}{System Prompt}@
You're a psychologist and social scientist working on a project to identify whether a person has stress based on reddit posts.
given a set of reddit posts, we want to generate hypotheses that are useful for deciding people's stress status (has stress or no stress) based on reddit post.

Using the given relevant literatures, refine the hypothesis pairs provided.
The desired hypotheses should identify specific patterns that occur across the provided posts.

Each hypothesis should contain a pair of the following:
a. A hypothesis about what makes the post more likely to indicate that the poster has stress
b. The opposite hypothesis about what makes the post more likely to indicate that the poster does not have stress

Generate refined hypotheses in the format of 1. [hypothesis], 2. [hypothesis], ... @\textcolor{mypurple}{<num\_hypotheses>}@. [hypothesis].
The hypotheses should analyze what kind of posts are likely to indicate stress or no stress.

@\textcolor{mygray}{User Prompt}@
We have some key findings from a series of research papers that might be useful for generating hypotheses:
@\textcolor{mygray}{··· information from literature here ···}@
We have some hypotheses need to be refined:
@\textcolor{mygray}{... hypotheses to be refined here ...}@
Please refine these hypotheses to make them more specific and useful for deciding people's stress status (has stress or no stress) based on reddit post.
Generate refined hypotheses in the format of 1. [hypothesis], 2. [hypothesis], ... @\textcolor{mypurple}{<num\_hypotheses>}@. [hypothesis].
Refined hypotheses:
\end{lstlisting}

\begin{lstlisting}[caption={Multiple-Hypothesis-Based Inference.},label={lst:inference:dreaddit},firstnumber=auto,xleftmargin=0pt]
@\textcolor{mygray}{System Prompt}@
You're a psychologist and social scientist working on a project to identify whether a person has stress based on reddit posts.
From past experiences, you learned some patterns. 
You need to determine whether each of the patterns holds for the current reddit post, and also predict whether the poster of the reddit post has stress or not based on the content of the post.
Give an answer. The answer should be "has stress" or "no stress".
Give your final answer in the format of {Final answer: answer}

@\textcolor{mygray}{User Prompt}@
Our learned patterns: @\textcolor{mypurple}{<generated\_hypotheses>}@    
A reddit post is the following: @\textcolor{mypurple}{<post>}@

Given the pattern you learned above, give an answer of whether the poster of the reddit post has stress or not based on the content of the post.
Think step by step.
First step: Think about which pattern can be applied to the reddit post.
Second step: Based on the pattern, does the poster of a reddit post has stress or not? Answer should be "has stress" or "no stress".
Final step: give your final answer in the format of {Final answer: answer}
\end{lstlisting}

\subsection{Persuasive Argument Prediction}
\label{appendix:prompts:persuasive_pairs}

\begin{lstlisting}[caption={Data-Based Hypothesis Generation with HypoGeniC.},label={lst:hypogenic:persuasion},firstnumber=auto,xleftmargin=0pt]
@\textcolor{mygray}{System Prompt}@
You are an intelligent rhetorician and debater who masters persuasiveness in language.
Given a pair of arguments, you are asked to determine which one of them uses more persuasive language. The two arguments are often on the same topic and are similar, so focus on their differences.
What difference between the two arguments makes one more persuasive than the other?
You will be given a set of observations of the format:
Argument 1: [argument_1]
Argument 2: [argument_2]
Observation: The first/second argument uses more persuasive language.
Based on the observations, please generate hypotheses that are useful for explaining why one argument uses more persuasive language than the other.
These hypotheses should identify patterns, phrases, wordings etc. that occur across the provided examples. They should also be generalizable to new instances.
Please propose @\textcolor{mypurple}{<num\_hypotheses>}@ possible hypotheses and generate them in the format of 1. [hypothesis], 2. [hypothesis], ... @\textcolor{mypurple}{<num\_hypotheses>}@. [hypothesis].

@\textcolor{mygray}{User Prompt}@
Here are the Observations:
@\textcolor{mygray}{··· more examples here ···}@

Please generate hypotheses that can help determine which argument uses more persuasive language.
Please propose @\textcolor{mypurple}{<num\_hypotheses>}@ possible hypotheses.

Generate them in the format of 1. [hypothesis], 2. [hypothesis], ... @\textcolor{mypurple}{<num\_hypotheses>}@. [hypothesis]. 

Proposed hypotheses:
\end{lstlisting}

\begin{lstlisting}[caption={Literature-Based Hypothesis Generation.},label={lst:litgen:persuasion},firstnumber=auto,xleftmargin=0pt]
@\textcolor{mygray}{System Prompt}@
You are an intelligent rhetorician and debater who masters persuasiveness in language.
Given a pair of arguments, you are asked to determine which one of them uses more persuasive language. The two arguments are often on the same topic and are similar, so focus on their differences.
What difference between the two arguments makes one more persuasive than the other?
You will be given a set of literature of the format:
Title: [title]
Key Findings: [summary]
Based on the literature, please generate hypotheses that are useful for explaining why one argument uses more persuasive language than the other.
These hypotheses should identify patterns, phrases, wordings etc. that you can find in the literature. They should also be generalizable to new instances.
Please propose @\textcolor{mypurple}{<num\_hypotheses>}@ refined hypotheses and generate them in the format of 1. [hypothesis], 2. [hypothesis], ... @\textcolor{mypurple}{<num\_hypotheses>}@. [hypothesis].

@\textcolor{mygray}{User Prompt}@
Here are some key findings from a series of research papers that might be useful for generating hypotheses:
@\textcolor{mygray}{··· information from literature here ···}@

Please generate hypotheses that can help determine which argument uses more persuasive language.
Please propose @\textcolor{mypurple}{<num\_hypotheses>}@ possible hypotheses.

Generate them in the format of 1. [hypothesis], 2. [hypothesis], ... @\textcolor{mypurple}{<num\_hypotheses>}@. [hypothesis]. 

Proposed hypotheses:
\end{lstlisting}

\begin{lstlisting}[caption={Paper Summarization.},label={lst:summarize:persuasion},firstnumber=auto,xleftmargin=0pt]
@\textcolor{mygray}{System Prompt}@
You are a helpful assistant for summarizing key findings in research papers on a given topic.

@\textcolor{mygray}{User Prompt}@
Summarize the following research paper, focusing ONLY on this question: What characterizes texts that use more persuasive language? In other words, how can one determine which one of two sentences uses more persuasive language?
Focus on hypotheses of what characterizes texts that use more persuasive language, do not include technical details in the paper. 
@\textcolor{mygray}{... literature texts here ...}@
\end{lstlisting}

\begin{lstlisting}[caption={Hypothesis Refinement Based on Data.},label={lst:data_refine:persuasion},firstnumber=auto,xleftmargin=0pt]
@\textcolor{mygray}{System Prompt}@
You are an intelligent rhetorician and debater who masters persuasiveness in language.
Given a pair of arguments, you are asked to determine which one of them uses more persuasive language. The two arguments are often on the same topic and are similar, so focus on their differences.
What difference between the two arguments makes one more persuasive than the other?
You will be given a set of observations of the format:
Argument 1: [argument_1]
Argument 2: [argument_2]
Observation: The first/second argument uses more persuasive language.
Based on the observations, please refine hypotheses provided to make them more useful for explaining why one argument uses more persuasive language than the other.
These hypotheses should identify patterns, phrases, wordings etc. that occur across the provided examples. They should also be generalizable to new instances.
Please propose @\textcolor{mypurple}{<num\_hypotheses>}@ refined hypotheses and generate them in the format of 1. [hypothesis], 2. [hypothesis], ... @\textcolor{mypurple}{<num\_hypotheses>}@. [hypothesis].

@\textcolor{mygray}{User Prompt}@
Here are the Observations:
@\textcolor{mygray}{··· more examples here ···}@

And here are the previous hypotheses:
@\textcolor{mygray}{... hypotheses to be refined here ...}@

Please generate refined hypotheses that can help determine which argument uses more persuasive language.
Please propose @\textcolor{mypurple}{<num\_hypotheses>}@ refined hypotheses.

Generate them in the format of 1. [hypothesis], 2. [hypothesis], ... @\textcolor{mypurple}{<num\_hypotheses>}@. [hypothesis]. 

Refined hypotheses:
\end{lstlisting}

\begin{lstlisting}[caption={Hypothesis Refinement Based on Literature.},label={lst:lit_refine:persuasion},firstnumber=auto,xleftmargin=0pt]
@\textcolor{mygray}{System Prompt}@
You are an intelligent rhetorician and debater who masters persuasiveness in language.
Given a pair of arguments, you are asked to determine which one of them uses more persuasive language. The two arguments are often on the same topic and are similar, so focus on their differences.
What difference between the two arguments makes one more persuasive than the other?
You will be given a set of literature of the format:
@\textcolor{mygray}{··· information from literature here ···}@
Based on the literature, please refine hypotheses provided to make them more useful for explaining why one argument uses more persuasive language than the other.
These hypotheses should identify patterns, phrases, wordings etc. that you can find in the literature. They should also be generalizable to new instances.
Please propose @\textcolor{mypurple}{<num\_hypotheses>}@ refined hypotheses and generate them in the format of 1. [hypothesis], 2. [hypothesis], ... @\textcolor{mypurple}{<num\_hypotheses>}@. [hypothesis].

@\textcolor{mygray}{User Prompt}@
Here are some key findings from a series of research papers that might be useful for generating hypotheses:
@\textcolor{mygray}{··· information from literature here ···}@

And here are the previous hypotheses:
@\textcolor{mygray}{... hypotheses to be refined here ...}@

Please generate refined hypotheses that can help determine which argument uses more persuasive language.
Please propose @\textcolor{mypurple}{<num\_hypotheses>}@ refined hypotheses.

Generate them in the format of 1. [hypothesis], 2. [hypothesis], ... @\textcolor{mypurple}{<num\_hypotheses>}@. [hypothesis]. 

Refined hypotheses:
\end{lstlisting}

\begin{lstlisting}[caption={Multiple-Hypothesis-Based Inference.},label={lst:inference:persuasion},firstnumber=auto,xleftmargin=0pt]
@\textcolor{mygray}{System Prompt}@
You are an intelligent rhetorician and debater who masters persuasiveness in language.
Given a pair of arguments, you are asked to determine which one of them uses more persuasive language. The two arguments are often on the same topic and are similar, so focus on their differences.
From past experiences, you learned some patterns.
Now, at each time, you should apply the learned patterns to a new pair of arguments and determine which one uses more persuasive language.
The answer for the more persuasive language should be of the form "the _ argument" where _ is either first or second. 
Please give your final answer in the format of {Final answer: the _ argument uses more persuasive language}

@\textcolor{mygray}{User Prompt}@
Our learned patterns: @\textcolor{mypurple}{<generated\_hypotheses>}@    
Given the patterns you learned above, determine which of the following arguments uses more persuasive language:
Argument 1: @\textcolor{mypurple}{<first\_argument>}@
Argument 2: @\textcolor{mypurple}{<second\_argument>}@

Think step by step.
Step 1: Think about which learned patterns can be applied to the arguments.
Step 2: Analyze the difference between "Argument 1" and "Argument 2".
Step 3: Based on the pattern, which argument uses more persuasive language?
You MUST give your final answer in the following format:
Final answer: the _ argument uses more persuasive language.
\end{lstlisting}

\section{Automated Experiments Implementation Details}
\label{appendix:implementation_details}

\subsection{Partitioning of IND and OOD datasets}
\label{appendix:implementation_details:OOD_parititioning}
\paragraph{Deception Detection} As stated in Section 4.1, our IND datasets are from \deceptive, which contains 800 truthful hotel reviews from the web and 800 deceptive reviews gathered from Mechanical Turk \citep{ott-etal-2013-negative}. The OOD dataset, \fourcities \citep{oodhotelreviews} consists of 640 hotel reviews from four cities and different web sources following the same procedure as \deceptive.

\paragraph{AI-Generated Content (AIGC) Detection} As discussed in Section 4.1, our AIGC Detection task consists of two subtasks, GPTGC and LlamaGC. The IND dataset for GPTGC contains GPT generated stories and human-written stories, while the one for LlamaGC includes Llama-generated stories and human-written stories. The OOD dataset of GPTGC is the IND dataset of LlamaGC and vice versa.

\paragraph{Mental Stress Detection} IND and OOD datasets are separated based on the source subreddits, or topic-specific communities, from which the Reddit posts are collected. Instances of the IND dataset are from ptsd, anxiety, and domestic violence subreddits, while those of OOD dataset are from relationships and homeless subreddits.

\paragraph{Persuasive Argument Prediction} IND and OOD datasets are partitioned according to the source corpora of the non-LLM-generated texts. Examples from IND datasets are from ElecDeb60to20 \citep{goffredo-etal-2023-argument}, Persuasion For Good \citep{wang2020persuasiongoodpersonalizedpersuasive}, and Webis-Clickbait-17 \citep{potthast-etal-2018-crowdsourcing}, while OOD dataset is from PT-Corpus \citep{da-san-martino-etal-2019-fine}.

\subsection{Specificity Boost}
We further observed that sometimes the solely literature-based hypotheses generated by gpt-4o-mini are often too short and brief, making it harder to apply during inference.
To address this, we add a LLM-based specificity booster after the literature-based hypothesis generation that adds more concrete illustrations and examples to each of the hypotheses based solely on its pre-training knowledge. Specifically we apply the specificity booster on our Deception Detection, Mental Stress Detection, and Persuasive Argument Prediction tasks. The specificity booster is not applied to Llama-3.1-70B-Instruct because it can already generate reasonably specific hypotheses. 

\subsection{Refinement and Union Implementation}
\label{appendix:implementation_details:refine_union_implementation}
\paragraph{Refinement} is implemented as an extension based on the original HypoGeniC pipeline. During the initialization stage, an LLM $\calM_G$ is instructed to generate an initial hypothesis bank $\calH_{\calL+\calD}^0$ based on a set of initial examples $\calD_{\operatorname{init}}$ and a series of generated paper summaries $\calS$, i.e., $\calH_{\calL+\calD}^0=\calM_G(\calS,\calD_{\operatorname{init}})$. These initial hypotheses are then evaluated and re-ranked using the same reward function as in HypoGeniC. 
In the update stage at time $t$, if the size of the wrong examples bank $\calW$ reaches $w_{\operatorname{max}}$, a set of new hypotheses is generated by feeding both the wrong examples bank and paper summaries to $\calM_G$. $\calH_{\calL+\calD}^t$ is then updated with the new hypothesis according to the reward, following the same procedure as HypoGeniC.

\paragraph{Union and Redundancy Elimination} is implemented by combining the hypothesis bank generated using \hypogenic $\calH_\calD$ or \refinemethod $\calH_{\calL+\calD}$and the bank generated by our literature-based hypothesis generation method. We first generate the two hypothesis banks separately using \hypogenic, \refinemethod, and \paperonly, following the procedures described above and in Section 3. Each hypothesis bank is then fed to a redundancy checker module. For a hypothesis bank of size 20, the LLM-based redundancy checker checks each pair of hypotheses and see if one entails the other, with results recorded as a $20 \times 20$ matrix $\mathcal{A}$ of 1 (redundant) or 0 (not redundant). To create the new no-redundancy hypothesis bank $\calH_{\operatorname{new}}$, we first rank the hypotheses based on their training accuracy. Each time we take the best-performing hypothesis $h$ out of the original hypothesis bank $\calH$ and check if there exists a hypothesis $h_{\operatorname{new}}$ in $\calH_{\operatorname{new}}$ such that redundancy is recorded in $\mathcal{A}$ for the pair $h$ and $h_{\operatorname{new}}$, i.e., $\mathcal{A}_{h,h_{\operatorname{new}}} = 1$ or $\mathcal{A}_{h_{\operatorname{new}},h} = 1$. If yes, $h$ is moved out of the original bank $\calH$ and skipped; if not, $h$ is moved to $\calH_{\operatorname{new}}$ with a rank determined by its training accuracy. 

After removing redundancies of hypothesis banks, we unite two hypothesis banks to create a final bank $\calH_{\operatorname{final}}$ with a balanced prioritization strategy. We first move the top 10 hypotheses from the \hypogenic or \refinemethod hypothesis bank to $\calH_{\operatorname{final}}$. If there is less than 10 hypotheses in the banks, we move all hypotheses to $\calH_{\operatorname{final}}$. Then we randomly choose hypotheses from the literature-based hypothesis bank until the size of $\calH_{\operatorname{final}}$ reaches 20.

\subsection{Multiple-Hypothesis Inference Implementation}
\label{appendix:implementation_details:multi_hyp_inference_implementation}
During multiple-hypothesis based inference, each time we feed a LLM with our final hypothesis bank $\calH$ of size 20 (see Appendix B.5) and an instance $(x,\_)$ of our IND or OOD datasets with labels removed. The LLM $\calM_I$ is asked to generate an answer for the given instance using Chain-of-Thought prompting \citep{weiChainThoughtPrompting2022} that considers both the relevance of the hypotheses to the given instance and the utility of the hypothesis bank (see Appendix A for the exact prompts we used). The prediction is denoted as $\hat{y}=\calM_I(\calH,x)$. We then compute the average accuracy for all data instances in the held-out IND and OOD sets with the model predictions. For F1 scores, we report the macro-averaged F1 scores.

\subsection{Technical Details of NotebookLM and HyperWrite}
\label{appendix:implementation_details:notebooklm_hyperwrite}
NotebookLM is an LLM-powered research assistance tool that generates source-grounding responses to user prompts. Specifically in our case, collected literature are uploaded in the NotebookLM interface, followed by a hypothesis generation prompt asking to generate hypotheses based on given literature. Given its functionality and our usage, it is placed under the literature-based hypothesis generation category in our evaluations.

For HyperWrite, we use its Hypothesis Maker function, which is an AI-driven tool that generates hypotheses based on a given research question.  Though there is no publicly available technical report for this tool, it generally leverages LLM's pre-training knowledge and literature information to produce hypotheses.

\subsection{Hyperparameters}
\label{appendix:implementation_details:hyperparameters}
We use the same set of hyperparameters across all tasks, models, and methods.

During the training stage of HypoGeniC, the limit of the hypothesis bank size, $H_{\operatorname{max}}$, is set to 20, and the size of training set is set to 200. In the initialization stage, we set $\mathtt{num\_init}=10$. In the update stage, we use reward coefficient $\alpha=0.5$, $w_{\operatorname{max}}=10$, $k=10$, and generate 10 hypothesis per update. 

In our \refinemethod method, the round of refinement $N_{\operatorname{refine}}$ is set to 6.

We use 5 random seeds for multiple-hypothesis inference: 11376, 8271, 39660, 543, 3. 

Across all tasks and methods and for both GPT-4o-mini and Llama-3.1-70B-Instruct, we use $\mathtt{temperature}=1 \times 10^{-5}$ and $\mathtt{max\_tokens}=4000$.

\subsection{Licensing Details}
\label{appendix:implementation_details:licensing_details}
\deceptive is released under CC BY-NC-SA 3.0, and \persuasion is released under CC BY-NC 4.0. The \writingprompts dataset which we use to create the AIGC Detection datasets are under MIT License. The \llamagc and \gptgc datasets will be released under the same licensing as this work, CC BY 4.0 License, should it be accepted. \dreaddit and \fourcities do not have licenses specified in their original papers, but are considered under CC BY 4.0 and CC BY-NC-SA 3.0 license respectively as they are ACL materials.

For the LLMs, \shortgpt is a proprietary and not released under any open-source license, while \shortllama is released under Llama 3.1 Community License Agreement. 

Throughout our study, we find that we are in compliance with the licensing agreements of all the datasets and models used in this work.

\subsection{Estimated Cost}
For \shortllama, we run all of our experiments with 4 NVIDIA A100s, and it takes on average 1.5 hours to run all of our hypothesis generation pipelines, including \hypogenic, \refinemethod, \unionhypogenic, and \unionhyporefine. With \shortgpt, the average cost for running the same pipelines is \$0.6.

\section{Human Study Details}
\label{appendix:human_study}

\subsection{Decision-making Utility Study Details}
\label{appendix:human_study:practical_relevance}

\begin{table*}[t]
\centering
\resizebox{1\textwidth}{!}{%
\begin{tabular}{p{12cm}c}
\hline
\textbf{Hypotheses} & \textbf{Frequency of Selection} \\
\hline
\hline
\textbf{Hypothesis 1:} Reviews that present a balanced perspective by detailing both positive and negative experiences with specific examples (e.g., “the room was spacious and clean, but the noise from the street was disruptive at night”) are more likely to be truthful, whereas reviews that express extreme sentiments without acknowledging any redeeming qualities (e.g., “everything was perfect” or “it was a total disaster”) are more likely to be deceptive. & 50.00\% \\
\textbf{Hypothesis 2:} Reviews that mention specific dates of stay or unique circumstances surrounding the visit (e.g., “We stayed during the busy Memorial Day weekend and faced long lines”) are more likely to be truthful, while reviews that use vague temporal references (e.g., “I stayed recently”) without concrete details are more likely to be deceptive, as they often lack the specificity that suggests a real and engaged experience. & 34.44\% \\
\textbf{Hypothesis 3:} Reviews that provide detailed sensory descriptions of the hotel experience, such as the specific decor of the room, the quality of bedding, and the overall ambiance (e.g., “the room featured luxurious furnishings, high-thread-count sheets, and soft lighting that created a relaxing atmosphere”) are more likely to be truthful, while reviews that use vague or overly simplistic descriptors (e.g., “the hotel was nice and comfortable”) are more likely to be deceptive. & 46.39\% \\
\textbf{No hypothesis selected} & 7.50\% \\
\hline
\end{tabular}
}
\caption{How often humans use hypotheses in Deception Detection human study. We allow users to select multiple hypotheses for each instance they make prediction on, so the total frequency can exceed 100\%.}
\label{tab:utility-deception-detection}
\end{table*}

\begin{table*}[t]
\centering
\resizebox{1\textwidth}{!}{%
\small
\begin{tabular}{@{}l p{0.75\textwidth}@{}}
\toprule
\textbf{Criteria} & \textbf{Texts}\\

\midrule
\multirow{5}{*}{\textbf{Clarity}} 
    & 1. Highly ambiguous (The hypothesis is presented in a highly ambiguous manner, lacking clear definition and leaving significant room for interpretation or confusion.)\\
    & 2. Somewhat clear but vague (The hypothesis is somewhat defined but suffers from vague terms and insufficient detail, making it challenging to grasp its meaning or how it could be tested.)\\
    & 3. Moderately clear (The hypothesis is stated in a straightforward manner, but lacks the depth or specificity needed to fully convey its nuances, assumptions, or boundaries.)\\
    & 4. Clear and precise (The hypothesis is clearly articulated with precise terminology and sufficient detail, providing a solid understanding of its assumptions and boundaries with minimal ambiguity.)\\
    & 5. Exceptionally clear (The hypothesis is exceptionally clear, concise, and specific, with every term and aspect well-defined, leaving no room for misinterpretation and fully encapsulating its assumptions, scope, and testability.)\\
\midrule
\multirow{5}{*}{\textbf{Novelty}} 
    & 1. Not novel (The hypothesis has already been shown, proven, or is widely known, closely mirroring existing ideas without introducing any new perspectives.)\\
    & 2. Minimally novel (The hypothesis shows slight novelty, introducing minor variations or nuances that build upon known ideas but do not offer significant new insights.)\\
    & 3. Moderately novel (The hypothesis demonstrates moderate novelty, presenting some new perspectives or angles that provide meaningful, but not groundbreaking, avenues for exploration.)\\
    & 4. Notably novel (The hypothesis is notably novel, offering unique nuances or perspectives that are well-differentiated from existing ideas, representing valuable and fresh contributions to the field.)\\
    & 5. Highly novel (The hypothesis is highly novel, introducing a pioneering perspective or idea that has not been previously explored, opening entirely new directions for future research.)\\
\midrule
\multirow{5}{*}{\textbf{Plausibility}} 
    & 1. Not plausible (The hypothesis does not make sense at all, lacking logical or empirical grounding and failing to align with established knowledge or principles.)\\
    & 2. Minimally plausible (The hypothesis has significant plausibility challenges, making sense in limited contexts but contradicting existing evidence or lacking coherence with established theories.)\\
    & 3. Moderately plausible (The hypothesis makes sense overall and aligns with general principles or existing knowledge but has notable gaps or uncertainties that raise questions about its validity.)\\
    & 4. Mostly plausible (The hypothesis is mostly plausible, grounded in logical reasoning and existing evidence, with only minor uncertainties or assumptions that could reasonably be addressed.)\\
    & 5. Highly plausible (The hypothesis is highly plausible, fully aligning with established knowledge and logical reasoning, will likely be supported in experiments or theoretical consistency, and highly likely to be true.)\\
\bottomrule
\end{tabular}
}
\caption{Criteria used for human evaluation of the generated hypotheses.
}
\label{tab:likert-scales}
\end{table*}

\begin{figure*}[t]
    \centering
    \includegraphics[width=0.95\textwidth]{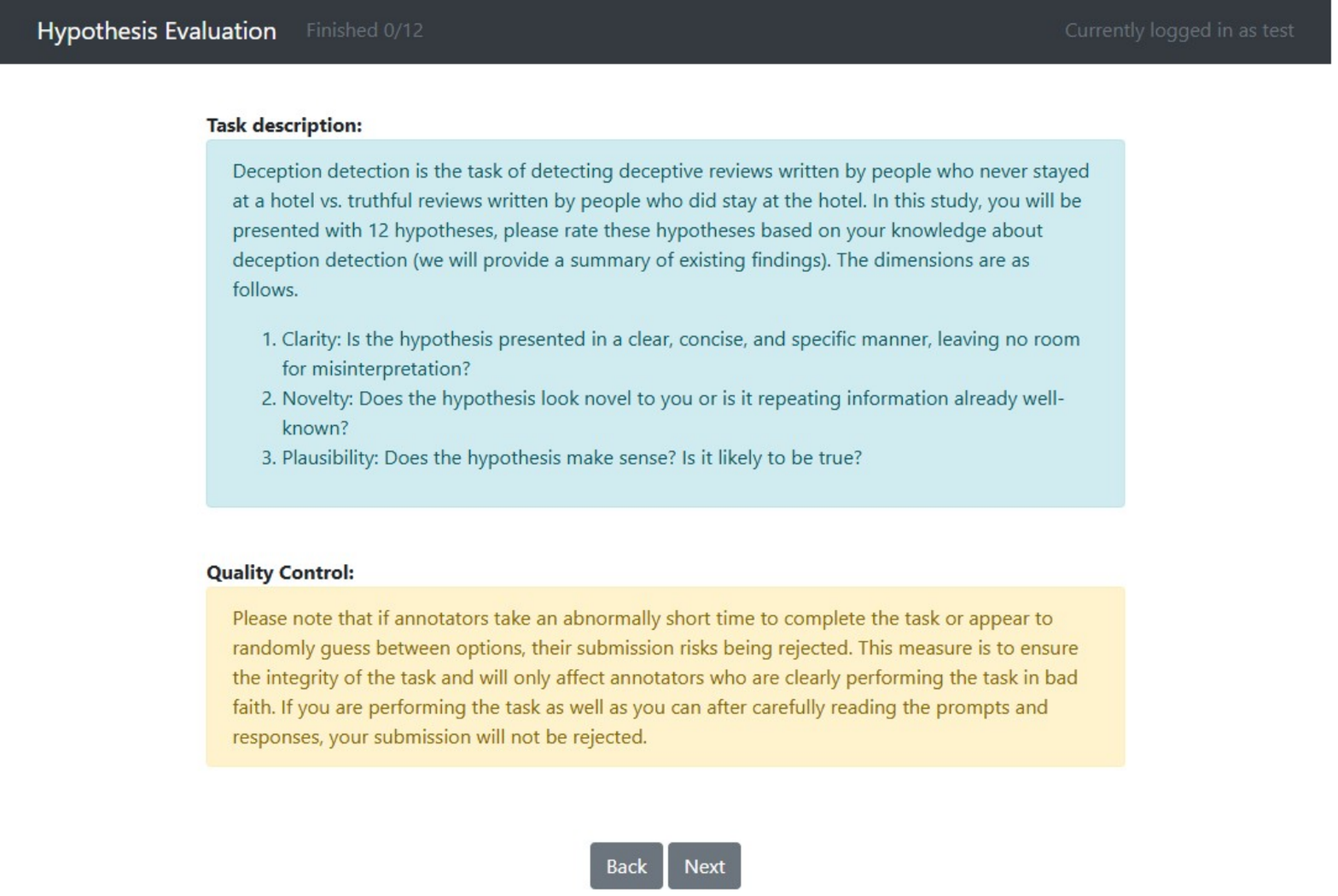}
    \caption{
    Instruction page for likert rating.
    }
    \label{fig:likert-instruction}
\end{figure*}

\begin{figure*}[t]
    \centering
    \includegraphics[width=0.9\textwidth]{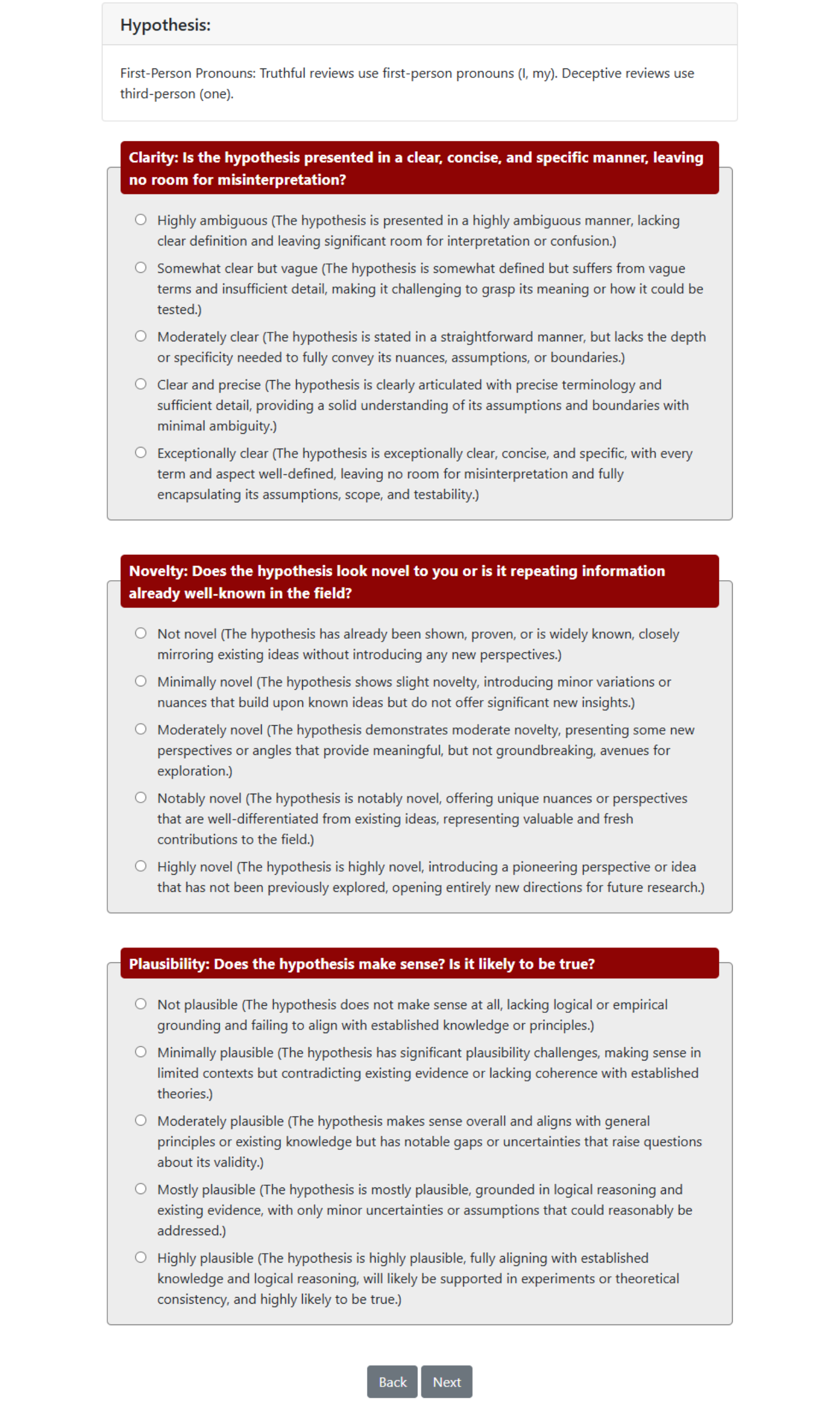}
    \caption{
    Annotation page for likert rating.
    }
    \label{fig:likert-annotation}
\end{figure*}

\begin{figure*}[t]
    \centering
    \includegraphics[width=0.95\textwidth]{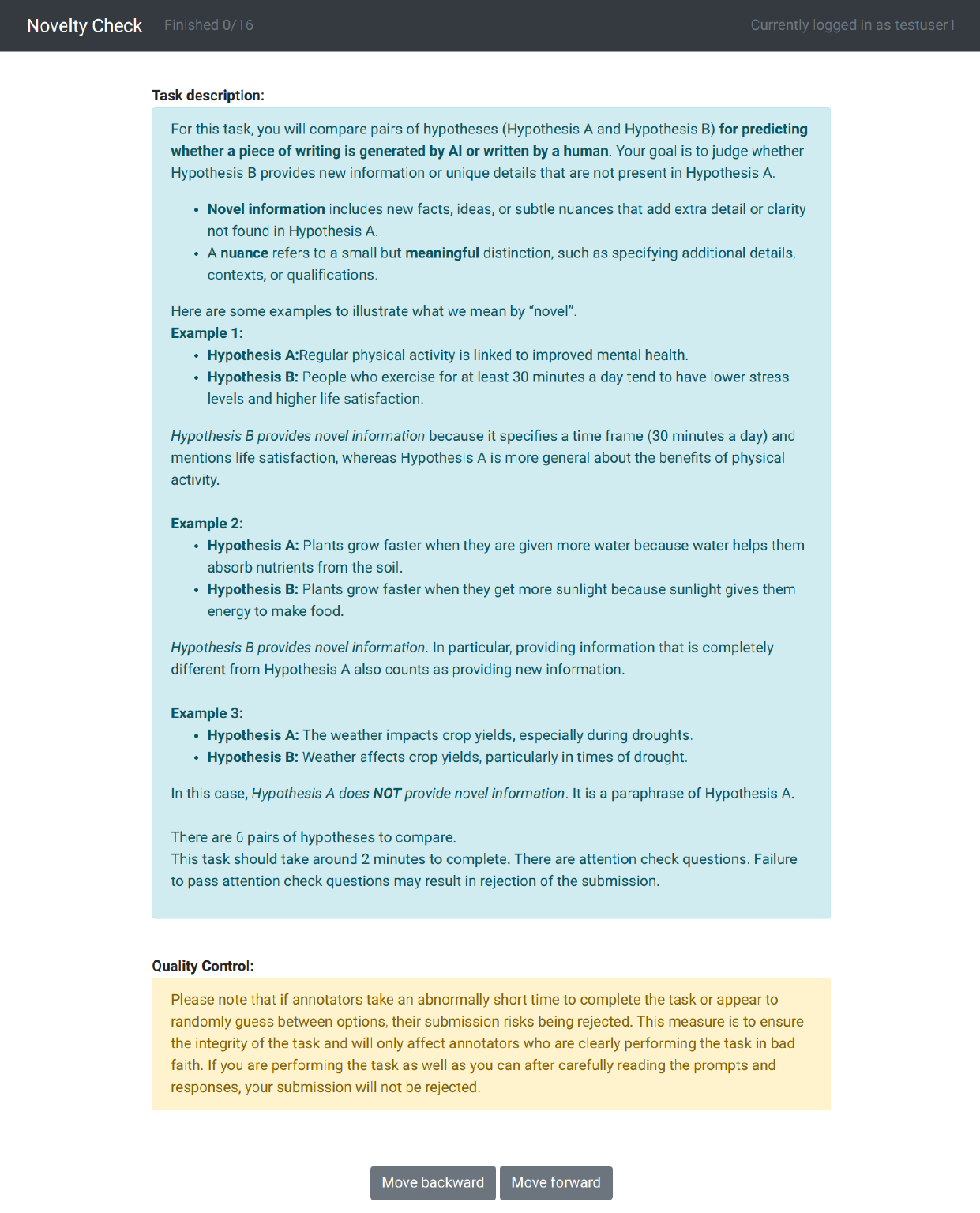}
    \caption{
    Instruction page for novelty check.
    }
    \label{fig:novelty-aigc-instruction}
\end{figure*}

\begin{figure*}[t]
    \centering
    \includegraphics[width=0.95\textwidth]{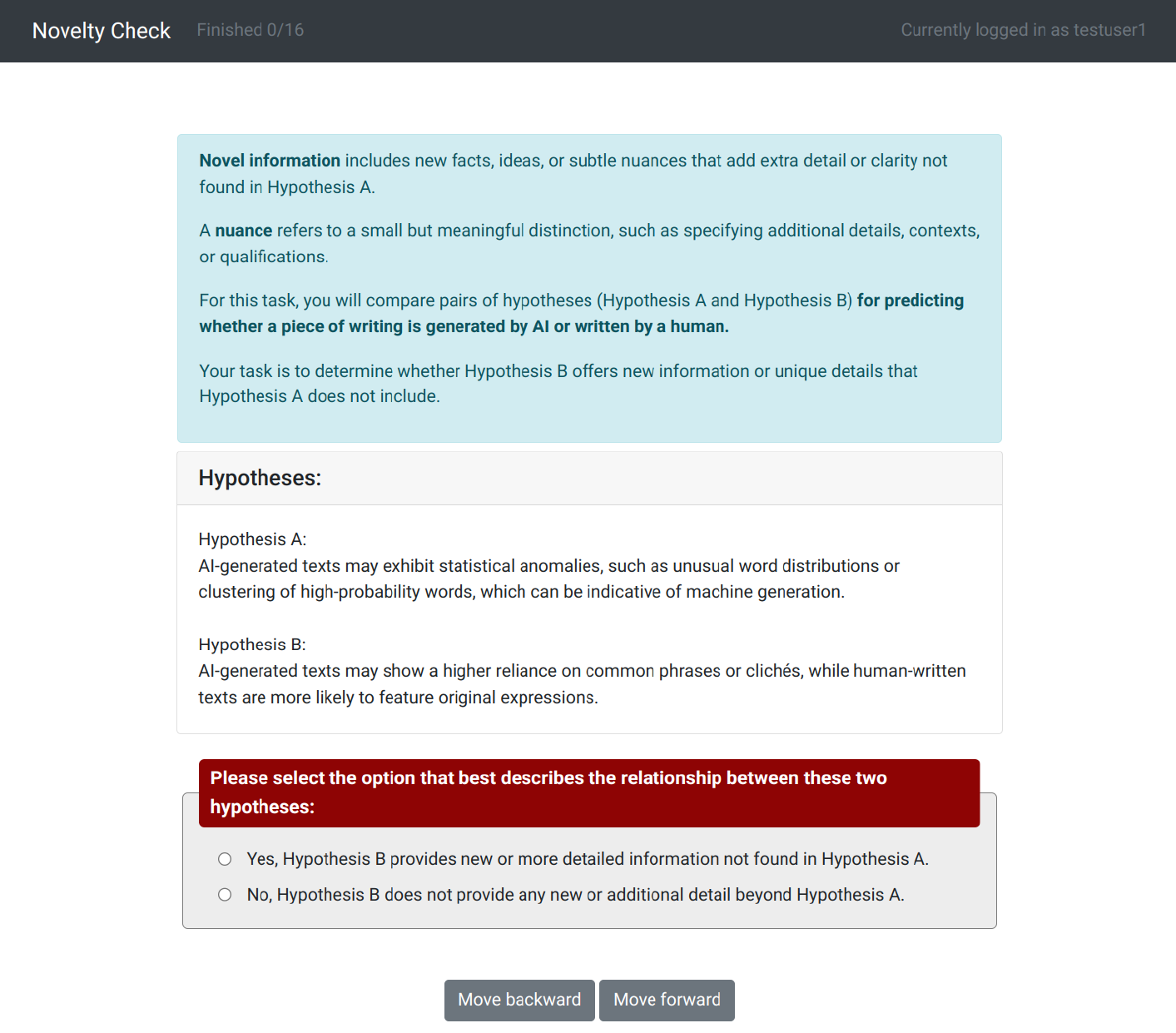}
    \caption{
    Annotation page for novelty check. 
    }
    \label{fig:novelty-aigc-annotation}
\end{figure*}

\begin{figure*}[t]
    \centering
    \includegraphics[width=0.95\textwidth]{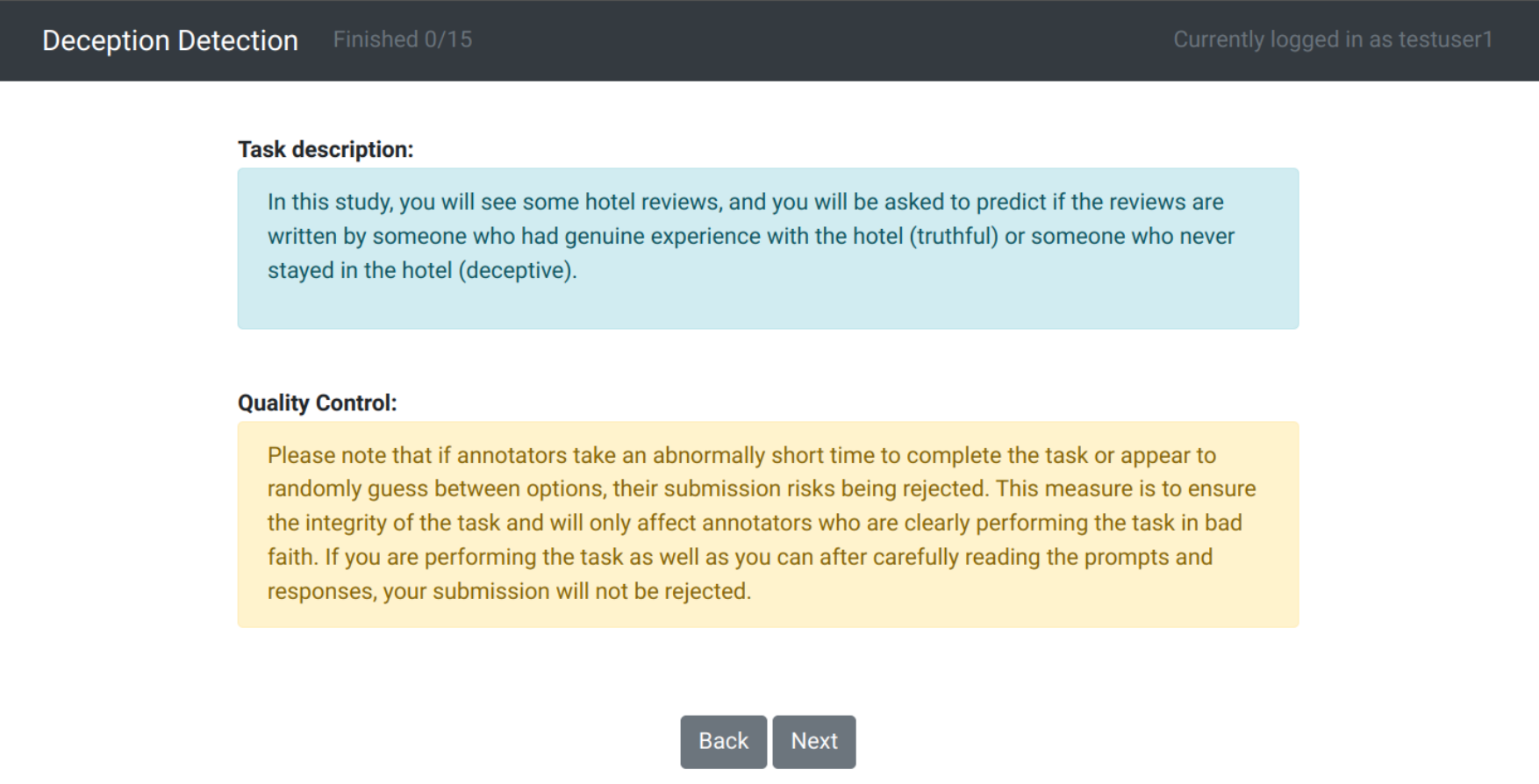}
    \caption{
    Instruction page for prediction task without hypotheses. 
    }
    \label{fig:regular-hotel-instruction}
\end{figure*}

\begin{figure*}[t]
    \centering
    \includegraphics[width=0.95\textwidth]{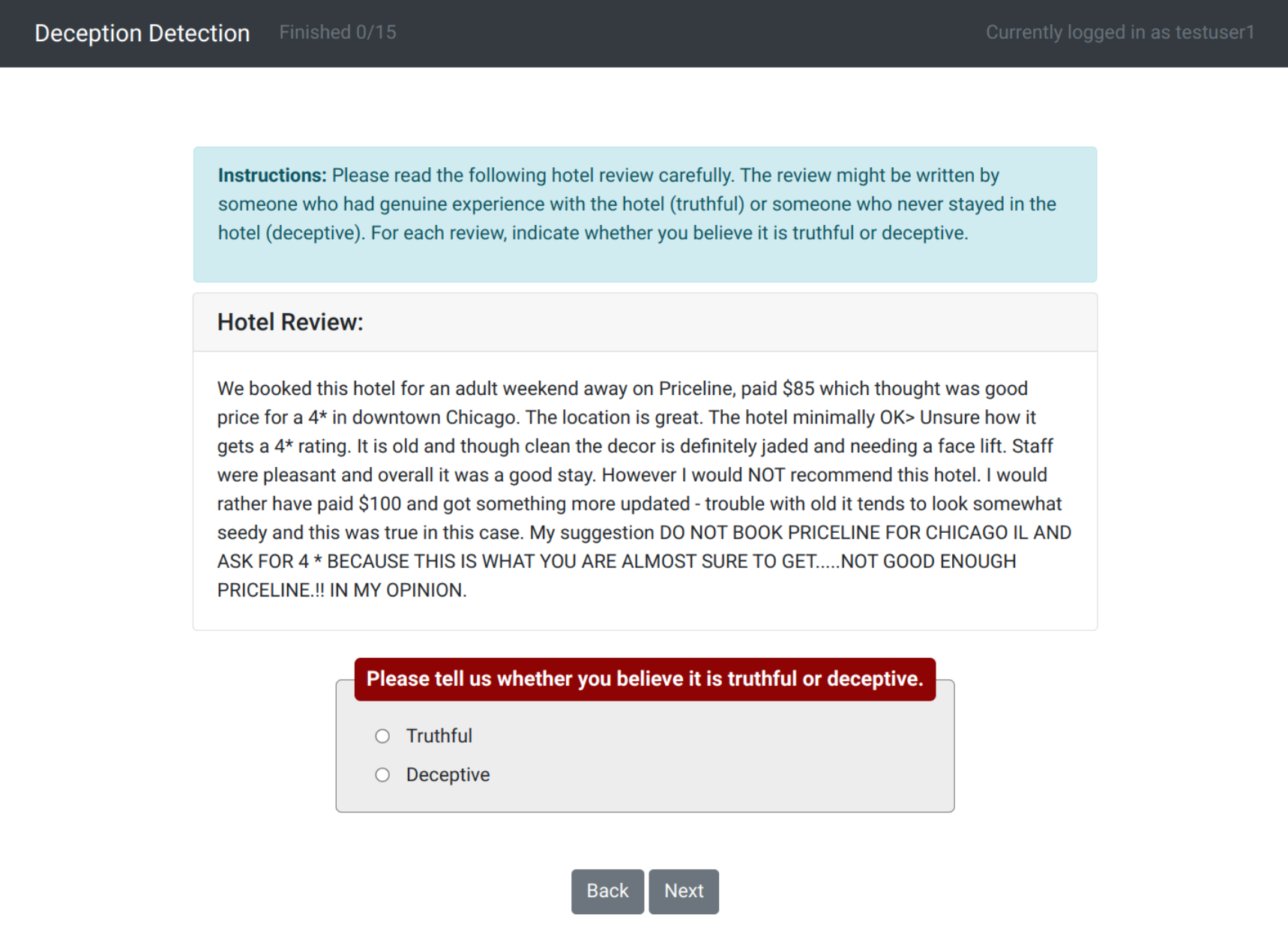}
    \caption{
    Annotation page for prediction task without hypotheses. 
    }
    \label{fig:regular-hotel-annotation}
\end{figure*}

\begin{figure*}[t]
    \centering
    \includegraphics[width=0.95\textwidth]{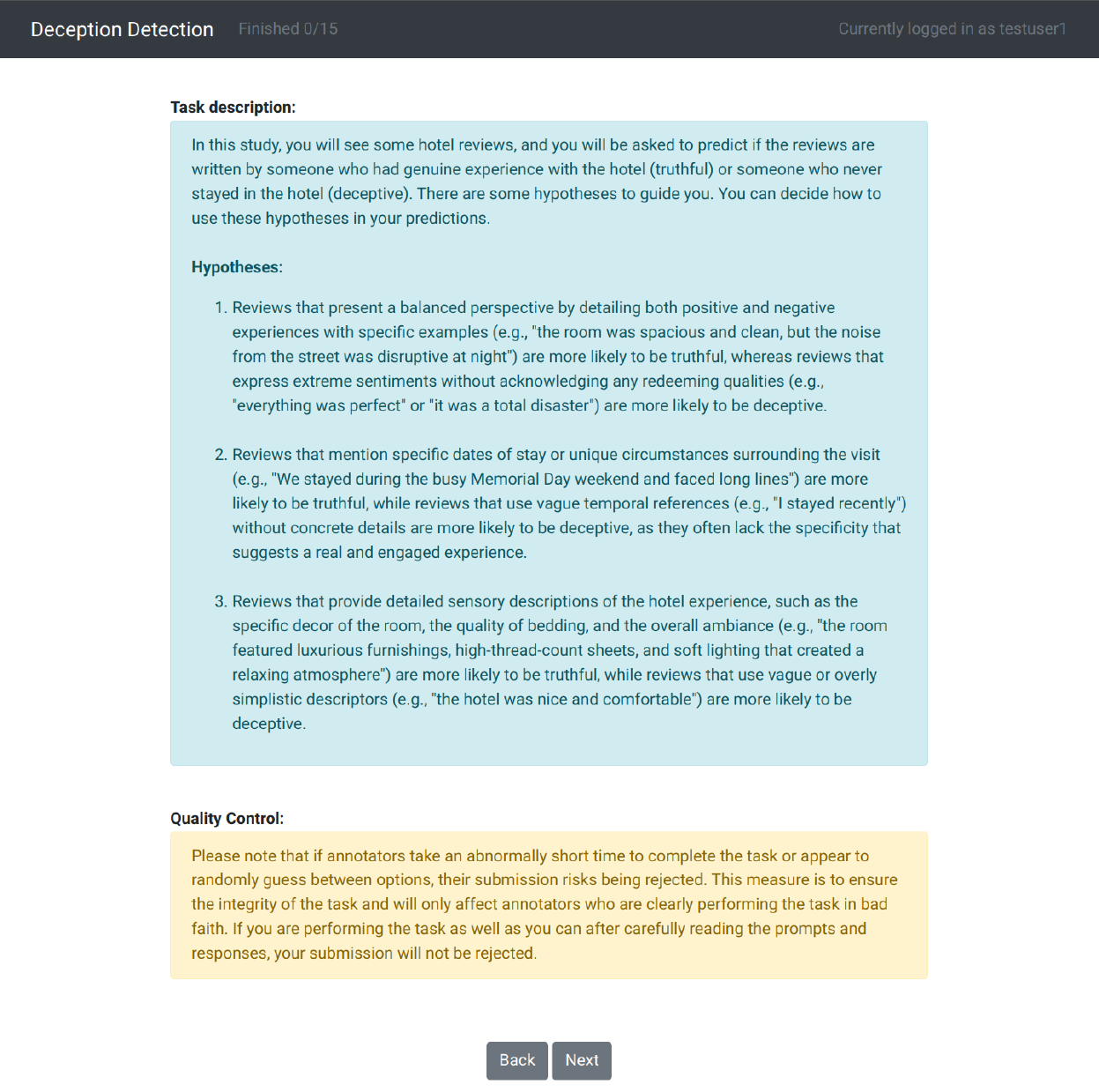}
    \caption{
    Instruction page for prediction task with the guide of hypotheses. 
    }
    \label{fig:guided-hotel-instruction}
\end{figure*}

\begin{figure*}[t]
    \centering
    \includegraphics[width=0.95\textwidth]{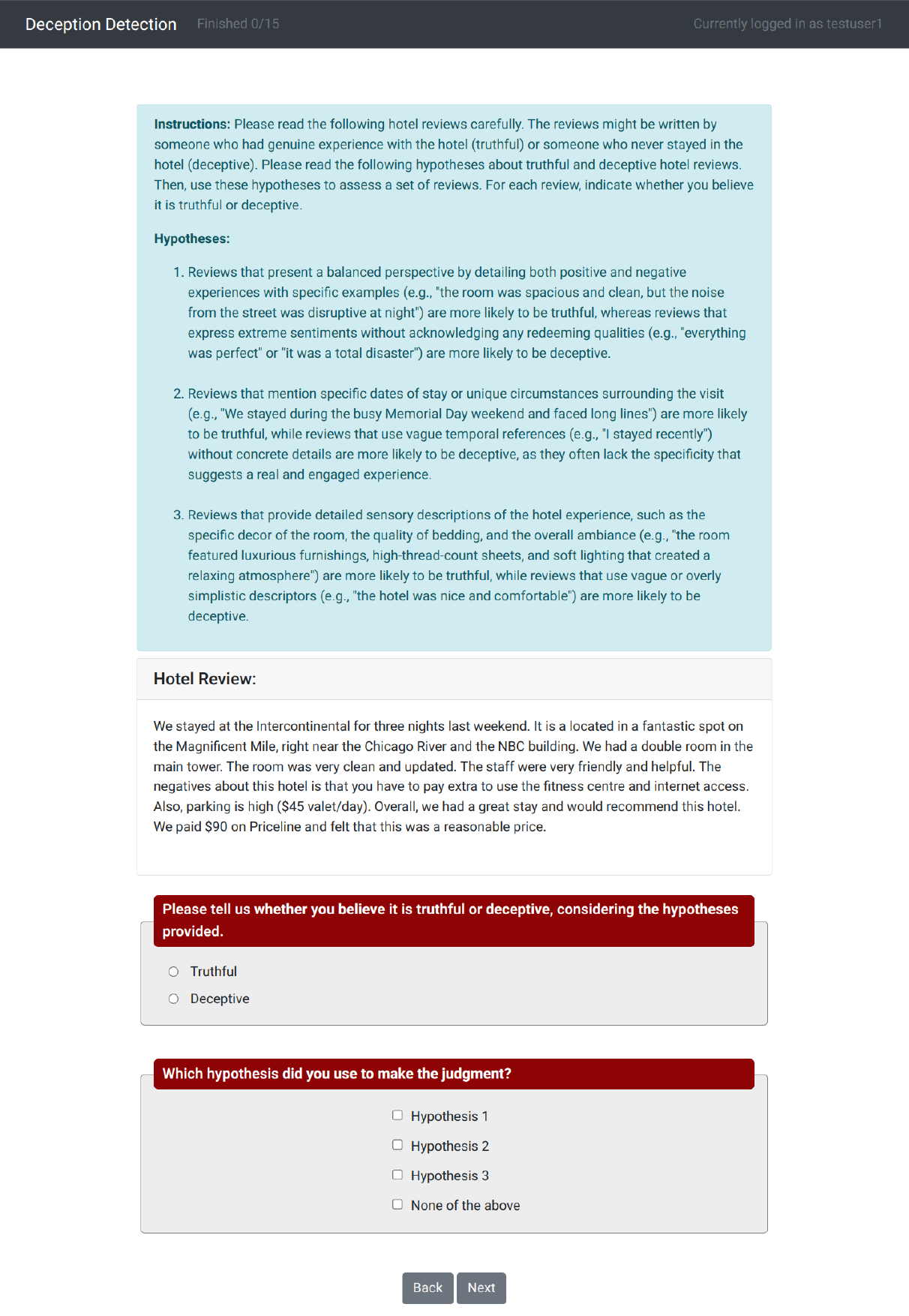}
    \caption{
    Annotation page for prediction task with the guide of hypotheses. 
    }
    \label{fig:guided-hotel-annotation}
\end{figure*}

The instructions of the practical relevance study can be found in \cref{fig:regular-hotel-instruction} and \cref{fig:guided-hotel-instruction}. For the interface, we present an example of the control group interface for Deception Detection in \cref{fig:regular-hotel-annotation}, and examples of the experiment group interface in \cref{fig:guided-hotel-annotation}.

The subjects of the control group are instructed to perform deception detection or AIGC (\gptgc) detection tasks without any assistance from the hypotheses. Subjects in the experiment group are asked to first read the presented 3 hypotheses and then make their predictions on the given instance. They are then required to choose which ones, if any, of the hypotheses that were used in their prediction. At the end of the study, participants in the experiment group are also asked to give overall rating and assessment of the helpfulness of the given hypotheses. There are five scales: ``Not at all helpful'', ``Slightly helpful'', ``Moderately helpful'', ``Very helpful'', and ``Extremely helpful''.

We choose top 3 hypotheses from the hypothesis bank generated using \unionhyporefine that cause the greatest drop in performance when removed from the hypotheses pool during multi-hypothesis inference. The chosen hypotheses for Deception Detection and AIGC Detection can be found in \cref{tab:utility-aigc-detection} and \cref{tab:utility-deception-detection}. 

We recruit 30 participants for the control group and 30 for the experimental group. 
For the control group, 4 people timed out, and 25 out of the remaining 26 participants passed attention checks. 
For the experimental group, 3 people timed out, and 22 out of the remaining 27 passed attention checks. 
We compute human accuracy based on responses from people who finished tasks in time and passed attention checks.
The average time spent is around 25 minutes and participants are timed out by the system if they spend more than 60 minutes in the study, which can happen when they accidentally leave the study website tab open but forget to do the task.

\subsection{Likert Rating Survey Details}
\label{appendix:human_study:likert}
We provide summaries of existing findings in Deception Detection to annotators and ask them to rate the hypotheses in terms of clarity, novelty, and plausibility. Each metric has five scales, which we include in \cref{tab:likert-scales}.
In particular, we manually select five representative hypotheses from the set of hypotheses generated by the different methods. We report the average human ratings in \cref{tab:likert-results}.
The instructions can be found in figure \cref{fig:likert-instruction}, and the interface for annotation can be found in \cref{fig:likert-annotation}.

\subsection{Novelty and Nuance Study Details}
\label{appendix:human_study:novelty}
For the Novelty and Nuance Study, we present the instructions for AIGC Detection in \cref{fig:novelty-aigc-instruction}. 
We showcase the interfaces for AIGC Detection in \cref{fig:novelty-aigc-annotation}.

For both Deception Detection and AIGC Detection, the two hypothesis banks compared are generated using \paperonly and \hypogenic respectively. 

We recruit 10 participants each task and all particpants passed attention the check question.

\subsection{IRB}
\label{appendix:human_study:irb}

We received IRB exempt (and will provide study number in the non-anonymous version of the paper).
For both of the human studies, we present a detailed description of the study, incentives, risks and benefits, confidentiality, and contacts \& questions in our consent form. The study proceeds only if the participant agrees to give consent.

\section{Additional Experiments}
\label{appendix:additional_experiments}
\begin{table*}[t]
\centering
\resizebox{1\textwidth}{!}{%
\begin{tabular}{@{}llcccccccccccc@{}}
\toprule
Model & Methods & \multicolumn{2}{c}{\deceptive OOD} & \multicolumn{2}{c}{\llamagc OOD} & \multicolumn{2}{c}{\gptgc OOD} & \multicolumn{2}{c}{\persuasion OOD} & \multicolumn{2}{c}{\dreaddit OOD} \\
\cmidrule(lr){3-4} \cmidrule(lr){5-6} \cmidrule(lr){7-8} \cmidrule(lr){9-10} \cmidrule(lr){11-12}
& & Accuracy & F1 & Accuracy & F1 & Accuracy & F1 & Accuracy & F1 & Accuracy & F1 \\
\midrule
\multirow{13}{*}{\shortstack{\textsc{GPT-4} \\ \textsc{MINI}}} 
    & \multicolumn{11}{c}{\textbf{No hypothesis}} \\
    & Zero-shot & 55.47 & 45.77 & 50.00 & 35.03 & 56.33 & 47.15 & 81.24 & 80.90 & 64.60 & 59.91 \\
    & Few-shot k=3 & 65.56 & 64.01 & 51.11 & 43.37 & 64.22 & 61.55 & 83.64 & 83.36 & 75.00 & 73.37 \\
    \cmidrule{2-12}
    
    & Zero-shot generation & 68.69 & 68.39 & 49.00 & 34.54 & 53.00 & 41.15 & 86.08 & 85.99 & 65.00 & 60.58 \\
    \cmidrule{2-12}
    
    & \multicolumn{11}{c}{\textbf{Literature-based}} \\
    & \paperonly & 59.22 & 57.31 & 49.00 & 33.45 & 54.00 & 41.65 & 78.80 & 78.62 & 67.68 & 64.52 \\
    & \hyperwrite & 61.63 & 57.97 & 49.67 & 33.76 & 52.67 & 39.96 & 82.36 & 82.09 & 68.76 & 65.92 \\
    & \notebooklm & 53.03 & 49.12 & 49.33 & 33.04 & 51.67 & 37.96 & 68.96 & 67.50 & 62.28 & 56.41 \\
    \cmidrule{2-12}
    
    & \multicolumn{11}{c}{\textbf{Data-driven}} \\
    & \hypogenic & 75.22 & 75.14 & 81.67 & 81.61 & 68.56 & 67.62 & 82.20 & 81.71 & 76.56 & 75.71 \\
    \cmidrule{2-12}
    & \multicolumn{11}{c}{\textbf{Literature + Data (This work)}}\\
    & \refinemethod  & \bf 77.78 & \bf 77.71 & 55.33 & 45.77 & 63.33 & 62.20 & 89.04 & 89.02 & 78.04 & 77.28 \\
    & Literature $\cup$ \hypogenic & 72.41 & 71.62 & \bf 83.00 & \bf 82.96 & \bf 69.22 & \bf 68.39 & \bf 89.88 & \bf 89.87 & 78.20 & 77.52 \\
    & Literature $\cup$ \refinemethod & 77.19 & 77.17 & 55.33 & 45.77 & 63.00 & 61.81 & 89.52 & 89.51 & \bf 79.24 & \bf 78.61 \\
\midrule

\multirow{13}{*}{\shortstack{\textsc{Llama} \\ 70B-I}} 

    & \multicolumn{11}{c}{\textbf{No hypothesis}}\\
    & Zero-shot & 62.87 & 58.45 & 58.67 & 50.79 & 63.00 & 57.61 & 85.60 & 85.59 & 64.56 & 59.98 \\
    & Few-shot k=3 & 68.56 & 67.25 & 70.45 & 67.53 & 76.00 & 74.97 & 86.80 & 86.76 & 69.44 & 66.47 \\
    \cmidrule{2-12}
    
    & Zero-shot generation & 56.28 & 40.27 & 50.67 & 35.90 & 55.67 & 45.61 & 88.16 & 88.13 & 66.16 & 62.59 \\
    \cmidrule{2-12}
    
    & \multicolumn{11}{c}{\textbf{Literature-based}}\\
    & \paperonly & 64.25 & 53.97 & 50.00 & 33.33 & 49.67 & 33.76 & 80.56 & 80.51 & 66.04 & 61.93 \\
    & \hyperwrite & 58.62 & 31.37 & 50.67 & 35.36 & 54.00 & 42.10 & 83.24 & 83.10 & 74.40 & 73.12 \\
    & \notebooklm & 57.81 & 36.91 & 49.33 & 33.61 & 50.67 & 35.90 & 67.64 & 66.41 & 66.56 & 62.83 \\
    \cmidrule{2-12}
    
    & \multicolumn{11}{c}{\textbf{Data-driven}}\\
    & \hypogenic & 62.06 & 56.89 & 78.67 & 78.53 & 78.00 & 77.26 & 88.44 & 88.38 & 75.48 & 74.55 \\
    \cmidrule{2-12}
    
    & \multicolumn{11}{c}{\textbf{Literature + Data (This work)}}\\
    & \refinemethod & 72.16 & 71.85 & 67.00 & 66.37 & 66.67 & 63.53 & 87.52 & 87.48 & \bf 78.92 & \bf 78.55 \\
    & Literature $\cup$ \hypogenic & \bf 73.72 & \bf 73.02 & \bf 81.33 & \bf 81.19 & \bf 78.67 & \bf 78.06 & 86.72 & 86.64 & 72.56 & 70.78 \\
    & Literature $\cup$ \refinemethod  & 71.75 & 71.33 & 66.67 & 65.79 & 65.67 & 62.67 & \bf 88.76 & \bf 88.73 & 74.80 & 73.55 \\
\bottomrule
\end{tabular}
}
\caption{Accuracy and F1 scores on the held-out OOD datasets. Literature + data outperforms all other methods in every model and task configurations. The bolded numbers outperform the few-shot method with statistical significance, as determined by a paired t-test using five random seeds.
}
\label{tab:ood_table_full}
\end{table*}

\begin{table*}[t]
\centering
\resizebox{1\textwidth}{!}{%
\begin{tabular}{@{}llcccccccccccc@{}}
\toprule
Model & Methods & \multicolumn{2}{c}{\deceptive IND} & \multicolumn{2}{c}{\llamagc IND} & \multicolumn{2}{c}{\gptgc IND} & \multicolumn{2}{c}{\persuasion IND} & \multicolumn{2}{c}{\dreaddit IND} \\
\cmidrule(lr){3-4} \cmidrule(lr){5-6} \cmidrule(lr){7-8} \cmidrule(lr){9-10} \cmidrule(lr){11-12}
& & Accuracy & F1 & Accuracy & F1 & Accuracy & F1 & Accuracy & F1 & Accuracy & F1 \\
\midrule
\multirow{13}{*}{\shortstack{\textsc{GPT-4} \\ \textsc{MINI}}} 
    & \multicolumn{11}{c}{\textbf{No hypothesis}} \\
    & Zero-shot & 56.56 & 51.66 & 56.33 & 47.15 & 50.00 & 35.03 & 83.72 & 83.59 & 62.32 & 56.24 \\
    & Few-shot k=3 & 62.60 & 61.40 & 63.67 & 61.54 & 54.78 & 49.50 & 85.24 & 85.14 & 67.48 & 63.59 \\
    \cmidrule{2-12}
    
    & Zero-shot generation & 60.16 & 60.15 & 54.33 & 44.36 & 49.67 & 33.18 & 87.72 & 87.71 & 62.24 & 56.11 \\
    \cmidrule{2-12}
    
    & \multicolumn{11}{c}{\textbf{Literature-based}} \\
    & \paperonly & 65.60 & 64.53 & 52.00 & 39.58 & 50.00 & 33.33 & 78.80 & 78.80 & 62.76 & 56.91 \\
    & \hyperwrite & 58.88 & 54.29 & 52.67 & 39.96 & 49.67 & 33.76 & 86.16 & 86.13 & 64.96 & 60.18 \\
    & \notebooklm & 50.40 & 48.79 & 51.67 & 37.96 & 49.33 & 33.04 & 75.92 & 75.78 & 60.52 & 53.37 \\
    \cmidrule{2-12}
    
    & \multicolumn{11}{c}{\textbf{Data-driven}} \\
    & \hypogenic & 68.32 & 67.60 & 72.33 & 71.70 & 66.33 & 64.88 & 84.68 & 84.52 & 70.40 & 67.90 \\
    \cmidrule{2-12}
    & \multicolumn{11}{c}{\textbf{Literature + Data (This work)}}\\
    & \refinemethod  & 68.56 & 68.43 & 62.33 & 57.08 & 57.33 & 54.41 & 89.76 & 89.75 & 70.76 & 68.31 \\
    & Literature $\cup$ \hypogenic & 68.76 & 67.66 & \bf 73.00 & \bf 72.49 & \bf 68.33 & \bf 67.54 & 89.84 & 89.84 & \bf 70.88 & \bf 68.46 \\
    & Literature $\cup$ \refinemethod & \bf 70.76 & \bf 70.68 & 60.33 & 54.30 & 55.00 & 51.65 & \bf 90.52 & \bf 90.52 & 69.88 & 67.21 \\
\midrule

\multirow{13}{*}{\shortstack{\textsc{Llama} \\ 70B-I}} 

    & \multicolumn{11}{c}{\textbf{No hypothesis}}\\
    & Zero-shot & 58.32 & 50.19 & 63.00 & 57.61 & 58.67 & 50.79 & 86.28 & 86.26 & 64.80 & 60.08 \\
    & Few-shot k=3 & 62.92 & 58.48 & 76.89 & 75.97 & 73.00 & 70.89 & 87.80 & 87.79 & 68.28 & 64.81 \\
    \cmidrule{2-12}
    
    & Zero-shot generation & 53.68 & 41.02 & 55.67 & 45.61 & 50.67 & 35.90 & 88.80 & 88.80 & 70.52 & 68.39 \\
    \cmidrule{2-12}
    
    & \multicolumn{11}{c}{\textbf{Literature-based}}\\
    & \paperonly & 62.96 & 58.39 & 51.33 & 36.23 & 49.67 & 22.61 & 80.32 & 80.32 & 66.08 & 62.09 \\
    & \hyperwrite & 52.28 & 26.75 & 54.00 & 42.10 & 50.67 & 35.36 & 87.48 & 87.46 & 71.12 & 69.24 \\
    & \notebooklm & 57.12 & 35.31 & 50.67 & 35.90 & 49.33 & 33.61 & 71.16 & 70.91 & 68.72 & 65.59 \\
    \cmidrule{2-12}
    
    & \multicolumn{11}{c}{\textbf{Data-driven}}\\
    & \hypogenic & 64.44 & 61.24 & \bf 78.33 & \bf 78.21 & \bf 80.67 & \bf 79.97 & \bf 91.24 & \bf 91.22 & 74.68 & 73.48 \\
    \cmidrule{2-12}
    
    & \multicolumn{11}{c}{\textbf{Literature + Data (This work)}}\\
    & \refinemethod & 71.68 & \bf 71.42 & 71.00 & 70.14 & 75.33 & 74.38 & 88.92 & 88.89 & \bf 78.68 & \bf 78.22 \\
    & Literature $\cup$ \hypogenic & 70.28 & 68.76 & \bf 78.33 & 78.13 & 80.00 & 79.22 & 89.36 & 89.33 & 70.68 & 68.28 \\
    & Literature $\cup$ \refinemethod  & \bf 72.60 & 67.80 & 71.00 & 70.14 & 74.00 & 72.84 & 90.88 & 90.88 & 73.20 & 71.60 \\
\bottomrule
\end{tabular}
}
\caption{Accuracy and F1 scores on the held-out IND datasets. Literature + data outperforms all other methods in 7 out of 10 configurations. For \shortllama on \gptgc, \llamagc, and \persuasion, \hypogenic performs the best. This is likely due to that the literature in these tasks do not offer helpful information for the IND data, but they can still provide useful information for the tasks in general. As in \cref{tab:ood_table_full}, our approaches with literature + data performs the best in all configurations for the OOD datasets.
}
\label{tab:ind_table_main}
\end{table*}

In this section, we include more analysis on the robustness of our hypothesis generation methods.
\subsection{Llama-3.1-8B-Instruct Results}
\label{appendix:additional_experiments_llama8b_results}
In \cref{tab:llama_8b_results}, we show the performance of Llama-3.1-8B-Instruct on the OOD and IND datasets for all tasks. We show that our approach with literature + data outperforms all other methods in 8 of the 10 total configurations. Across the 10 configurations, our method outperforms few-shot inference by 15.27\% on average, and it outperforms the best of literature-based methods and \hypogenic by 13.04\% and 4.88\%, respectively. This result further shows the effectiveness of our approach and provides evidence that our method can be applied with smaller models, highlighting its scalability and reproducibility.
\begin{table*}[t]
\centering
\resizebox{1\textwidth}{!}{%
\begin{tabular}{@{}llcccccccccccc@{}}
\toprule
 & Methods & \multicolumn{2}{c}{\deceptive} & \multicolumn{2}{c}{\llamagc} & \multicolumn{2}{c}{\gptgc} & \multicolumn{2}{c}{\persuasion} & \multicolumn{2}{c}{\dreaddit} \\
\cmidrule(lr){3-4} \cmidrule(lr){5-6} \cmidrule(lr){7-8} \cmidrule(lr){9-10} \cmidrule(lr){11-12}
& & Accuracy & F1 & Accuracy & F1 & Accuracy & F1 & Accuracy & F1 & Accuracy & F1 \\
\midrule
\multirow{13}{*}{\shortstack{OOD}} 
    & \multicolumn{11}{c}{\textbf{No hypothesis}} \\
    & Zero-shot & 48.25 & 41.37 & 48.00 & 32.43 & 52.00 & 40.48 & 63.96 & 47.15 & 62.80 & 57.40 \\
    & Few-shot k=3 & 54.00 & 53.09 & 46.20 & 33.78 & 53.20 & 44.35 & 77.96 & 77.79 & 63.32 & 57.81 \\
    \cmidrule{2-12}
    
    & Zero-shot generation & 25.25 & 20.97 & 48.53 & 32.68 & 59.00 & 53.76 & 76.88 & 76.84 & 60.56 & 54.68 \\
    \cmidrule{2-12}
    
    & \multicolumn{11}{c}{\textbf{Literature-based}} \\
    & \paperonly & 55.22 & 55.19 & 47.27 & 32.10 & 54.67 & 48.74 & 84.08 & 84.08 & 62.64 & 57.32 \\
    & \hyperwrite & 55.28 & 48.30 & 47.33 & 33.30 & 47.00 & 36.29 & 81.00 & 80.99 & 61.88 & 56.19 \\
    & \notebooklm & 52.09 & 47.01 & 49.00 & 32.89 & 50.67 & 36.42 & 58.84 & 58.83 & 62.20 & 56.89 \\
    \cmidrule{2-12}
    
    & \multicolumn{11}{c}{\textbf{Data-driven}} \\
    & \hypogenic & \bf 64.72 & \bf 64.00 & 63.53 & 60.48 & 70.33 & 68.97 & 78.96 & 78.89 & 68.64 & 66.18 \\
    \cmidrule{2-12}
    & \multicolumn{11}{c}{\textbf{Literature + Data (This work)}}\\
    & \refinemethod  & 56.91 & 51.33 & \bf 65.80 & \bf 64.92 & \bf 81.33 & \bf 81.14 & \bf 87.44 & \bf 87.44 & 68.20 & 67.85 \\
    & Literature $\cup$ \hypogenic & 54.34 & 44.51 & 63.93 & 61.00 & 70.60 & 69.21 & 83.84 & 83.76 & \bf 71.52 & \bf 69.75 \\
    & Literature $\cup$ \refinemethod & 61.91 & 61.34 & 65.73 & 64.68 & 81.07 & 80.85 & 85.96 & 85.93 & 68.80 & 67.28 \\
\midrule

\multirow{13}{*}{\shortstack{IND}} 

    & \multicolumn{11}{c}{\textbf{No hypothesis}}\\
    & Zero-shot & 50.88 & 41.34 & 52.26 & 40.98 & 48.00 & 32.43 & 70.92 & 49.05 & 63.76 & 58.39 \\
    & Few-shot k=3 & 58.20 & 54.86 & 51.80 & 43.89 & 47.33 & 33.49 & 71.84 & 71.55 & 63.20 & 57.81 \\
    \cmidrule{2-12}
    
    & Zero-shot generation & 28.64 & 22.60 & 49.20 & 35.16 & 50.27 & 39.36 & 78.00 & 77.79 & 61.60 & 55.51 \\
    \cmidrule{2-12}
    
    & \multicolumn{11}{c}{\textbf{Literature-based}}\\
    & \paperonly & 56.72 & 56.48 & 50.00 & 36.68 & 52.20 & 44.32 & 80.88 & 80.79 & 60.32 & 53.78 \\
    & \hyperwrite & 53.12 & 42.51 & 48.66 & 36.80 & 47.00 & 33.03 & 79.48 & 79.28 & 61.20 & 55.04 \\
    & \notebooklm & 50.84 & 44.57 & 50.47 & 36.52 & 48.67 & 32.74 & 63.72 & 63.47 & 63.40 & 58.47 \\
    \cmidrule{2-12}
    
    & \multicolumn{11}{c}{\textbf{Data-driven}}\\
    & \hypogenic & 57.60 & 54.29 & \bf 71.80 & \bf 70.61 & 70.07 & 68.66 & 79.16 & 79.06 & 66.12 & 62.53 \\
    \cmidrule{2-12}
    
    & \multicolumn{11}{c}{\textbf{Literature + Data (This work)}}\\
    & \refinemethod & 54.88 & 47.44 & 63.73 & 62.87 & 82.27 & 82.01 & 85.44 & 85.44 & \bf 69.92 & \bf 69.06 \\
    & Literature $\cup$ \hypogenic & 52.84 & 41.59 & 71.33 & 70.15 & 70.33 & 68.98 & 82.04 & 81.88 & 67.48 & 64.76 \\
    & Literature $\cup$ \refinemethod  & \bf 61.40 & \bf 58.89 & 64.13 & 63.30 & \bf 82.33 & \bf 82.09 & \bf 86.32 & \bf 86.32 & 67.56 & 65.35 \\
\bottomrule
\end{tabular}
}
\caption{Accuracy and F1 scores of Llama-3.1-8B-Instruct on the OOD and IND datasets. Literature + data outperforms all other methods in 4 out of 5 configurations for both OOD and IND datasets. This further validates the effectiveness of our methods even on smaller models.
}
\label{tab:llama_8b_results}
\end{table*}

\subsection{Robustness to Prompt Variations}
\label{appendix:additional_experiments_prompt_robustness}
Since our framework heavily relies on LLMs, we perform a robustness test of our hypothesis generation method with different prompt variations. Compared with the original prompts, we consider three prompt variations: modifying the hypothesis generation prompt for $\calM_G$, inference prompt for $\calM_I$, and both prompts. We show performance of all model and task configurations using these prompt variations in \cref{tab:prompt_sensitivity_ood} and \cref{tab:prompt_sensitivity_ind} for OOD and IND settings, respectively. For the OOD datasets, the accuracy only decreases by 0.20\% on average. For the 90 different configurations, 71 of them have a performance drop of less than 5\%, and 48 of them get an accuracy improvement. Additionally, with the IND datasets, the average accuracy gets an increase of 0.01\%. 74 out of the 90 configurations have a performance drop of less than 5\%, where 50 of them get an improvement. These additional results further illustrate the robustness of our method against variations of prompts.
\begin{table*}[t]
\centering
\resizebox{1\textwidth}{!}{%
\small
\begin{tabular}{llcccccccccccc}
\toprule
Model & Methods & \deceptive & \llamagc & \gptgc & \persuasion & \dreaddit \\

\midrule
\multirow{13}{*}{\shortstack{\textsc{GPT-4} \\ \textsc{MINI}}} 
    & \multicolumn{6}{c}{\textbf{\refinemethod}} \\
    & Original prompt & 77.78 & 55.33 & 63.33 & 89.04 & 78.04 \\
    & Prompt Variation 1 & 73.28 (\decrease 4.50) & 49.33 (\decrease 6.00) & 83.33 (\increase 20.00) & 91.80 (\increase 2.76) & 83.40 (\increase 5.36) \\
    & Prompt Variation 2 & 69.69 (\decrease 8.09) & 66.67 (\increase 11.34) & 69.33 (\increase 6.00) & 89.40 (\increase 0.36) & 75.60 (\decrease 2.44)\\
    & Prompt Variation 3 & 74.06 (\decrease 3.72) & 49.00 (\decrease 6.33) & 56.00 (\decrease 7.33) & 91.20 (\increase 2.16) & 71.20 (\decrease 6.84)\\
    \cmidrule{2-7}
    
    & \multicolumn{6}{c}{\textbf{Literature $\cup$ \hypogenic}} \\
    & Original prompt & 72.41 & 83.00 & 69.22 & 89.88 & 78.20 \\
    & Prompt Variation 1 & 74.53 (\increase 2.12) & 80.33 (\decrease 2.67) & 61.33 (\decrease 7.89) & 90.20 (\increase 0.32) & 74.20 (\decrease 4.00) \\
    & Prompt Variation 2 & 68.91 (\decrease 3.50) & 49.33 (\decrease 33.67) & 70.00 (\increase 0.78) & 91.60 (\increase 1.78) & 74.60 (\decrease 3.60)\\
    & Prompt Variation 3 & 73.75 (\increase 1.34) & 49.00 (\decrease 34.00) & 69.00 (\decrease 0.22) & 88.60 (\decrease 1.28) & 71.60 (\decrease 6.60)\\
    \cmidrule{2-7}
    
    & \multicolumn{6}{c}{\textbf{Literature $\cup$ \refinemethod}} \\
    & Original prompt & 77.19 & 55.33 & 63.00 & 89.52 & 79.24 \\
    & Prompt Variation 1 & 69.69 (\decrease 7.50) & 49.67 (\decrease 5.66) & 82.33 (\increase 19.33) & 90.40 (\increase 0.82) & 81.40 (\increase 2.16)\\
    & Prompt Variation 2 & 72.03 (\decrease 5.16) & 70.67 (\increase 15.34) & 63.67 (\increase 0.67) & 90.20 (\increase0.68) & 78.60 (\decrease 0.64) \\
    & Prompt Variation 3 & 69.38 (\decrease 7.81) & 49.00 (\decrease 6.33) & 57.33 (\decrease 5.67) & 90.00 (\increase 0.48) & 78.20 (\decrease 1.04) \\
\midrule

\multirow{13}{*}{\shortstack{\textsc{Llama} \\ 70B-I}} 

    & \multicolumn{6}{c}{\textbf{\refinemethod}} \\
    & Original prompt & 72.16 & 67.00 & 66.67 & 87.52 & 78.92 \\
    & Prompt Variation 1 & 63.13 (\decrease 9.03) & 66.00 (\decrease 1.00) & 68.33 (\increase 1.66) & 90.60 (\increase 3.08) & 81.60 (\increase 2.68) \\
    & Prompt Variation 2 & 70.47 (\decrease 1.69) & 81.00 (\increase 14.00) & 74.67 (\increase 8.00) & 84.20 (\decrease 3.32) & 70.80 (\decrease 8.12)\\
    & Prompt Variation 3 & 68.13 (\decrease 4.03) & 74.00 (\increase 7.00) & 74.00 (\increase 7.33) & 89.40 (\increase 1.88) & 81.40 (\increase 2.48) \\
    \cmidrule{2-7}
    
    & \multicolumn{6}{c}{\textbf{Literature $\cup$ \hypogenic}} \\
    & Original prompt & 73.72 & 81.33 & 78.67 & 86.72 & 72.56 \\
    & Prompt Variation 1 & 74.38 (\increase 0.66) & 70.67 (\decrease 10.66) & 80.33 (\increase 1.66) & 90.00 (\increase 3.28) & 75.60 (\increase 3.04) \\
    & Prompt Variation 2 & 72.97 (\decrease 0.75) & 78.33 (\decrease 3.00) & 85.33 (\increase 6.66) & 85.00 (\decrease 1.72) & 74.40 (\increase 1.84) \\
    & Prompt Variation 3 & 75.63 (\increase 1.91) & 80.00 (\decrease 1.33) & 83.67 (\increase 5.00) & 88.40 (\increase 1.68) & 75.00 (\increase 2.44) \\
    \cmidrule{2-7}
    
    & \multicolumn{6}{c}{\textbf{Literature $\cup$ \refinemethod}} \\
    & Original prompt & 71.75 & 66.67 & 65.67 & 88.76 & 74.80 \\
    & Prompt Variation 1 & 66.72 (\decrease 5.03) & 66.00 (\decrease 0.67) & 68.33 (\increase 2.66) & 89.00 (\increase 0.24) & 75.40 (\increase 0.60)\\
    & Prompt Variation 2 & 77.97 (\increase 6.22) & 81.00 (\increase 14.33) & 74.67 (\increase 9.00) & 88.80 (\increase 0.04) & 72.80 (\decrease 2.00) \\
    & Prompt Variation 3 & 69.53 (\decrease 2.22) & 74.00 (\increase 7.33) & 74.00 (\increase 8.33) & 89.60 (\increase 0.84) & 73.80 (\decrease 1.00) \\
\bottomrule
\end{tabular}
}
\caption{Accuracy numbers on OOD datasets with 4 different sets of prompts. The prompts used for the results in Table \ref{tab:ood_table_main}, Table \ref{tab:ood_table_full}, and Table \ref{tab:ind_table_main} are indicated with "original prompt". The prompt variations contain different paraphrases of the original prompts for hypothesis generation and hypothesis-based inference. Results show the robustness of our methods to different prompts.
}
\label{tab:prompt_sensitivity_ood}
\end{table*}

\begin{table*}[t]
\centering
\resizebox{1\textwidth}{!}{%
\small
\begin{tabular}{llcccccccccccc}
\toprule
Model & Methods & \deceptive & \llamagc & \gptgc & \persuasion & \dreaddit \\

\midrule
\multirow{13}{*}{\shortstack{\textsc{GPT-4} \\ \textsc{MINI}}} 
    & \multicolumn{6}{c}{\textbf{\refinemethod}} \\
    & Original prompt & 68.56 & 62.33 & 57.33 & 89.76 & 70.76 \\
    & Prompt Variation 1 & 65.40 (\decrease 3.16) & 52.00 (\decrease 10.33) & 84.67 (\increase 27.34) & 89.80 (\increase 0.04) & 73.20 (\increase 2.44) \\
    & Prompt Variation 2 & 67.00 (\decrease 1.56) & 68.33 (\increase 6.00) & 59.67 (\increase 2.34) & 88.80 (\decrease 0.96) & 69.80 (\decrease 0.96) \\
    & Prompt Variation 3 & 68.60 (\increase 0.04) & 53.33 (\decrease 9.00) & 53.00 (\decrease 4.33) & 88.60 (\decrease 1.16) & 69.40 (\decrease 1.36) \\
    \cmidrule{2-7}
    
    & \multicolumn{6}{c}{\textbf{Literature $\cup$ \hypogenic}} \\
    & Original prompt & 68.76 & 73.00 & 68.33 & 89.84 & 70.88 \\
    & Prompt Variation 1 & 69.20 (\increase 0.44) & 75.00 (\increase 2.00) & 54.67 (\decrease 13.66) & 88.40 (\decrease 1.44) & 65.80 (\decrease 5.08) \\
    & Prompt Variation 2 & 67.60 (\decrease 1.16) & 57.00 (\decrease 16.00) & 60.67 (\decrease 7.66) & 91.20 (\increase 1.36) & 69.00 (\decrease 1.88) \\
    & Prompt Variation 3 & 70.20 (\increase 1.44) & 54.67 (\decrease 18.33) & 59.33 (\decrease 9.00) & 87.20 (\decrease 2.64) & 66.80 (\decrease 4.08) \\
    \cmidrule{2-7}
    
    & \multicolumn{6}{c}{\textbf{Literature $\cup$ \refinemethod}} \\
    & Original prompt & 70.76 & 60.33 & 55.00 & 90.52 & 69.88 \\
    & Prompt Variation 1 & 64.00 (\decrease 6.76) & 51.67 (\decrease 8.66) & 85.67 (\increase 30.67) & 90.20 (\decrease 0.32) & 71.20 (\increase 1.32) \\
    & Prompt Variation 2 & 67.00 (\decrease 3.76) & 66.33 (\increase 6.00) & 54.67 (\decrease 0.33) & 90.80 (\increase 0.28) & 70.80 (\increase 0.92) \\
    & Prompt Variation 3 & 65.00 (\decrease 5.76) & 54.33 (\decrease 6.00) & 52.33 (\decrease 2.67) & 89.80 (\decrease 0.72) & 70.80 (\increase 0.92) \\
\midrule

\multirow{13}{*}{\shortstack{\textsc{Llama} \\ 70B-I}} 

    & \multicolumn{6}{c}{\textbf{\refinemethod}} \\
    & Original prompt & 71.68 & 71.00 & 75.33 & 88.92 & 78.68 \\
    & Prompt Variation 1 & 61.80 (\decrease 9.88) & 73.00 (\increase 2.00) & 80.00 (\increase 4.67) & 91.20 (\increase 2.28) & 80.80 (\increase 2.12) \\
    & Prompt Variation 2 & 70.80 (\decrease 0.88) & 77.00 (\increase 6.00) & 79.00 (\increase 3.67) & 89.80 (\increase 0.88) & 74.60 (\decrease 4.08) \\
    & Prompt Variation 3 & 70.00 (\decrease 1.68) & 79.00 (\increase 8.00) & 70.33 (\decrease 5.00) & 92.00 (\increase 3.08) & 80.00 (\increase 1.32) \\
    \cmidrule{2-7}
    
    & \multicolumn{6}{c}{\textbf{Literature $\cup$ \hypogenic}} \\
    & Original prompt & 70.28 & 78.33 & 80.00 & 89.36 & 70.68 \\
    & Prompt Variation 1 & 69.80 (\decrease 0.48) & 70.33 (\decrease 8.00) & 83.00 (\increase 3.00) & 91.80 (\increase 2.44) & 76.20 (\increase 5.52) \\
    & Prompt Variation 2 & 72.20 (\increase 1.92) & 79.33 (\increase 1.00) & 92.33 (\increase 12.33) & 89.60 (\increase 0.24) & 74.20 (\increase 3.52) \\
    & Prompt Variation 3 & 73.20 (\increase 2.92) & 81.00 (\increase 2.67) & 82.00 (\increase 2.00) & 92.40 (\increase 3.04) & 75.20 (\increase 4.52) \\
    \cmidrule{2-7}
    
    & \multicolumn{6}{c}{\textbf{Literature $\cup$ \refinemethod}} \\
    & Original prompt & 72.60 & 71.00 & 74.00 & 90.88 & 73.20 \\
    & Prompt Variation 1 & 64.80 (\decrease 7.80) & 73.00 (\increase 2.00) & 80.00 (\increase 6.00) & 89.80 (\decrease 1.08) & 76.20 (\increase 3.00) \\
    & Prompt Variation 2 & 75.20 (\increase 2.60) & 77.00 (\increase 6.00) & 74.67 (\increase 0.67) & 91.40 (\increase 0.52) & 74.00 (\increase 0.80) \\
    & Prompt Variation 3 & 67.20 (\decrease 5.40) & 79.00 (\increase 8.00) & 70.33 (\decrease 3.67) & 92.40 (\increase 1.52) & 77.60 (\increase 3.40) \\
\bottomrule
\end{tabular}
}
\caption{Accuracy numbers on IND datasets with 4 different sets of prompts.
}
\label{tab:prompt_sensitivity_ind}
\end{table*}

\subsection{Hyperparameter Search}
\label{appendix:additional_experiments_hyperparameter_search}
As introduced in \cref{sec:methods} and \cref{appendix:implementation_details:hyperparameters}, our hypothesis generation methods have some hyperparameters. Throughout all main experiments in \cref{sec:results}, we use the same set of hyperparameters, which is adopted from \hypogenic. As we show in \cref{sec:results}, this default choice of hyperparameters works consistently well across all different model and task configurations, highlighting the robustness of our framework. Here we conduct an additional hyperparameter search of $H_{\operatorname{max}}$. We show the results of using $H_{\operatorname{max}}=10,20,30$ in \cref{tab:hyperparameter_search_ood} and \cref{tab:hyperparameter_search_ind}, for the OOD and IND settings, respectively.

For the OOD datasets, changing $H_{\operatorname{max}}=10,20,30$ results in an average accuracy decrease of only 0.07\%. Out of the 60 different configurations, 51 of them get an accuracy drop of less than 5\%, where 31 of them get an increase. Moreover, with the IND datasets, different choices of $H_{\operatorname{max}}=10,20,30$ degrades average accuracy by 0.23\%. In 51 out of 60 cases, we get a performance drop of less than 5\%, and we get an improvement for 26 cases. These results suggest that although our default choice of hyperparameters may not be optimal for all tasks, our method is able to perform consistently well. This again highlights the robustness of our hypothesis generation framework with different hyperparameters.
\begin{table*}[t]
\centering
\resizebox{1\textwidth}{!}{%
\small
\begin{tabular}{llcccccccccccc}
\toprule
Model & Methods & \deceptive & \llamagc & \gptgc & \persuasion & \dreaddit \\

\midrule
\multirow{13}{*}{\shortstack{\textsc{GPT-4} \\ \textsc{MINI}}} 
    & \multicolumn{6}{c}{\textbf{\refinemethod}} \\
    & $H_{\operatorname{max}}$ = 10 & 78.75 (\increase 0.97) & 52.00 (\decrease 3.33) & 48.67 (\decrease 14.66) & 88.80 (\decrease 0.24) & 78.20 (\increase 0.16) \\
    & $H_{\operatorname{max}}$ = 20 & 77.78 & 55.33 & 63.33 & 89.04 & 78.04 \\
    & $H_{\operatorname{max}}$ = 30 & 79.69 (\increase 1.91) & 48.67 (\decrease 6.66) & 66.67 (\increase 3.34) & 90.40 (\increase 1.36) & 76.40 (\decrease 1.64) \\
    \cmidrule{2-7}
    
    & \multicolumn{6}{c}{\textbf{Literature $\cup$ \hypogenic}} \\
    & $H_{\operatorname{max}}$ = 10 & 73.00 (\decrease 0.59) & 68.00 (\decrease 15.00) & 60.33 (\decrease 8.89) & 90.60 (\increase 0.72) & 79.80 (\increase 1.60) \\
    & $H_{\operatorname{max}}$ = 20 & 72.41 & 83.00 & 69.22 & 89.88 & 78.20 \\
    & $H_{\operatorname{max}}$ = 30 & 74.60 (\increase 2.19) & 87.67 (\increase 4.67) & 82.33 (\increase 13.11) & 90.80 (\increase 0.92) & 75.40 (\decrease 2.80) \\
    \cmidrule{2-7}
    
    & \multicolumn{6}{c}{\textbf{Literature $\cup$ \refinemethod}} \\
    & $H_{\operatorname{max}}$ = 10 & 73.80 (\decrease 3.39) & 51.33 (\decrease 4.00) & 49.00 (\decrease 14.00) & 89.40 (\decrease 0.12) & 75.80 (\decrease 3.44) \\
    & $H_{\operatorname{max}}$ = 20 & 77.19 & 55.33 & 63.00 & 89.52 & 79.24 \\
    & $H_{\operatorname{max}}$ = 30 & 76.40 (\decrease 0.79) & 49.33 (\decrease 6.00) & 67.67 (\increase 4.67) & 90.80 (\increase 1.28) & 74.20 (\decrease 5.04) \\
\midrule

\multirow{13}{*}{\shortstack{\textsc{Llama} \\ 70B-I}} 

    & \multicolumn{6}{c}{\textbf{\refinemethod}} \\
    & $H_{\operatorname{max}}$ = 10 & 73.59 (\increase 1.43) & 71.33 (\increase 4.33) & 77.67 (\increase 10.00) & 84.00 (\decrease 3.52) & 79.00 (\increase 0.08) \\
    & $H_{\operatorname{max}}$ = 20 & 72.16 & 67.00 & 66.67 & 87.52 & 78.92 \\
    & $H_{\operatorname{max}}$ = 30 & 71.09 (\decrease 1.07) & 72.33 (\increase 5.33) & 78.33 (\increase 11.66) & 90.00 (\increase 2.48) & 72.80 (\decrease 6.12) \\
    \cmidrule{2-7}
    
    & \multicolumn{6}{c}{\textbf{Literature $\cup$ \hypogenic}} \\
    & $H_{\operatorname{max}}$ = 10 & 70.20 (\decrease 3.52) & 78.33 (\decrease 3.00) & 81.00 (\increase 2.33) & 86.80 (\increase 0.08) & 69.60 (\decrease 2.96) \\
    & $H_{\operatorname{max}}$ = 20 & 73.72 & 81.33 & 78.67 & 86.72 & 72.56 \\
    & $H_{\operatorname{max}}$ = 30 & 66.00 (\decrease 7.72) & 86.67 (\increase 5.34) & 81.00 (\increase 2.33) & 89.20 (\increase 2.48) & 75.80 (\increase 3.24) \\
    \cmidrule{2-7}
    
    & \multicolumn{6}{c}{\textbf{Literature $\cup$ \refinemethod}} \\
    & $H_{\operatorname{max}}$ = 10 & 69.20 (\decrease 2.55) & 71.33 (\increase 4.66) & 77.67 (\increase 12.00) & 87.20 (\decrease 1.56) & 77.80 (\increase 4.00) \\
    & $H_{\operatorname{max}}$ = 20 & 71.75 & 66.67 & 65.67 & 88.76 & 74.80 \\
    & $H_{\operatorname{max}}$ = 30 & 69.60 (\decrease 2.15) & 72.33 (\increase 5.66) & 78.33 (\increase 12.66) & 85.20 (\decrease 3.56) & 71.80 (\decrease 3.00) \\
\bottomrule
\end{tabular}
}
\caption{Accuracy numbers on OOD datasets with different limits on the hypothesis bank size $H_{\operatorname{max}}$.
}
\label{tab:hyperparameter_search_ood}
\end{table*}

\begin{table*}[t]
\centering
\resizebox{1\textwidth}{!}{%
\small
\begin{tabular}{llcccccccccccc}
\toprule
Model & Methods & \deceptive & \llamagc & \gptgc & \persuasion & \dreaddit \\

\midrule
\multirow{13}{*}{\shortstack{\textsc{GPT-4} \\ \textsc{MINI}}} 
    & \multicolumn{6}{c}{\textbf{\refinemethod}} \\
    & $H_{\operatorname{max}}$ = 10 & 65.60 (\decrease 2.94) & 53.67 (\decrease 8.66) & 49.67 (\decrease 7.66) & 89.00 (\decrease 0.76) & 69.80 (\decrease 0.96) \\
    & $H_{\operatorname{max}}$ = 20 & 68.56 & 62.33 & 57.33 & 89.76 & 70.76 \\
    & $H_{\operatorname{max}}$ = 30 & 67.40 (\decrease 1.16) & 59.33 (\decrease 3.00) & 66.00 (\increase 8.67) & 92.20 (\increase 2.44) & 70.80 (\increase 0.04) \\
    \cmidrule{2-7}
    
    & \multicolumn{6}{c}{\textbf{Literature $\cup$ \hypogenic}} \\
    & $H_{\operatorname{max}}$ = 10 & 69.60 (\increase 0.84) & 68.00 (\decrease 5.00) & 74.33 (\increase 6.00) & 90.20 (\increase 0.36) & 69.80 (\decrease 1.08) \\
    & $H_{\operatorname{max}}$ = 20 & 68.76 & 73.00 & 68.33 & 89.84 & 70.88 \\
    & $H_{\operatorname{max}}$ = 30 & 68.00 (\decrease 0.76) & 77.00 (\increase 5.00) & 86.67 (\increase 18.34) & 90.20 (\increase 0.36) & 66.00 (\decrease 4.88) \\
    \cmidrule{2-7}
    
    & \multicolumn{6}{c}{\textbf{Literature $\cup$ \refinemethod}} \\
    & $H_{\operatorname{max}}$ = 10 & 67.40 (\decrease 3.36) & 53.67 (\decrease 6.66) & 49.33 (\decrease 5.67) & 88.40 (\decrease 2.12) & 68.00 (\decrease 1.88) \\
    & $H_{\operatorname{max}}$ = 20 & 70.76 & 60.33 & 55.00 & 90.52 & 69.88 \\
    & $H_{\operatorname{max}}$ = 30 & 65.40 (\decrease 5.36) & 57.33 (\decrease 3.00) & 67.33 (\increase 12.33) & 90.40 (\decrease 0.12) & 68.00 (\decrease 1.88) \\
\midrule

\multirow{13}{*}{\shortstack{\textsc{Llama} \\ 70B-I}} 

    & \multicolumn{6}{c}{\textbf{\refinemethod}} \\
    & $H_{\operatorname{max}}$ = 10 & 66.80 (\decrease 4.88) & 72.33 (\increase 1.33) & 79.00 (\increase 3.67) & 87.80 (\decrease 1.12) & 79.20 (\increase 0.52) \\
    & $H_{\operatorname{max}}$ = 20 & 71.68 & 71.00 & 75.33 & 88.92 & 78.68 \\
    & $H_{\operatorname{max}}$ = 30 & 68.40 (\decrease 3.28) & 82.33 (\increase 11.33) & 77.33 (\increase 2.00) & 91.40 (\increase 2.48) & 71.20 (\decrease 7.48)\\
    \cmidrule{2-7}
    
    & \multicolumn{6}{c}{\textbf{Literature $\cup$ \hypogenic}} \\
    & $H_{\operatorname{max}}$ = 10 & 63.00 (\decrease 7.28) & 78.67 (\increase 0.34) & 85.33 (\increase 5.33) & 88.60 (\decrease 0.76) & 66.00 (\decrease 4.68) \\
    & $H_{\operatorname{max}}$ = 20 & 70.28 & 78.33 & 80.00 & 89.36 & 70.68 \\
    & $H_{\operatorname{max}}$ = 30 & 62.60 (\decrease 7.68) & 86.33 (\increase 8.00) & 80.33 (\increase 0.33) & 90.00 (\increase 0.64) & 74.60 (\increase 3.92) \\
    \cmidrule{2-7}
    
    & \multicolumn{6}{c}{\textbf{Literature $\cup$ \refinemethod}} \\
    & $H_{\operatorname{max}}$ = 10 & 65.40 (\decrease 8.20) & 72.33 (\increase 1.33) & 79.00 (\increase 5.00) & 89.80 (\decrease 1.08) & 72.20 (\decrease 1.00) \\
    & $H_{\operatorname{max}}$ = 20 & 72.60 & 71.00 & 74.00 & 90.88 & 73.20 \\
    & $H_{\operatorname{max}}$ = 30 & 68.40 (\decrease 4.20) & 72.33 (\increase 1.33) & 77.33 (\increase 3.33) & 88.20 (\decrease 2.68) & 70.80 (\decrease 2.40) \\
\bottomrule
\end{tabular}
}
\caption{Accuracy numbers on IND datasets with different limits on the hypothesis bank size $H_{\operatorname{max}}$.
}
\label{tab:hyperparameter_search_ind}
\end{table*}

\section{Examples of Generated Hypotheses and Qualitative Analysis}
\label{appendix:hypothesis_examples}

\subsection{Comparing Hypotheses from SciMON and Ours}
\label{appendix:hypothesis_examples:comparing_scimon}
In \cref{sec:related_work}, we briefly introduce the difference between research idea generation and our hypothesis generation work. To better illustrate this difference, we include a detailed comparison in \cref{tab:comparison_with_scimon}, consisting the generated research idea from SciMON \citep{wang2024scimon}, generated hypothesis with \refinemethod, and an existing finding from \citet{li2014generalrule} on Deception Detection. For the SciMON generated idea, we adopted from Table 11 in \citet{wang2024scimon}. These examples show that SciMON aims to generate ideas for a potential research project, where our method focus on generating possible explanations of a phenomenon. In addition, comparing with the existing finding from \citet{li2014generalrule}, our generated hypothesis is highly relevant to the field of interest, i.e., Deception Detection.
\begin{table*}[t]
    \centering
    \resizebox{0.95\textwidth}{!}{%
    \begin{tabular}{@{}lp{10cm}p{10cm}@{}}
    \toprule
    \textbf{Method} & \textbf{Example Hypotheses and Findings} \\ \midrule \midrule
    SciMON \citep{wang2024scimon} & Exploiting Social Media for Irish Language Learning: An Analysis of Twitter Data. In this context, we use social media data, particularly from Twitter, as a method for Irish
    language learning, because it provides a rich source of authentic and diverse language
    examples that can be used to enhance learning opportunities for L2 learners in a minority language setting. \\ 
    \midrule
    \refinemethod & Reviews that provide specific accounts of the checkin and check-out processes, including exact times, the names of staff members involved, and descriptions of any unique features or services utilized (e.g., "I used the self-check-in kiosk at 3 PM"), are more likely to be truthful. Conversely, reviews that mention issues like long wait times or check-in problems without contextual details or specific examples (e.g., "the check-in took too long") are more likely to be deceptive. \\ 
    \midrule
    \citet{li2014generalrule} & Deceptive reviews often contain a higher frequency of first-person singular pronouns, while truthful reviews may use these pronouns less frequently.\\
    \bottomrule
    \end{tabular}
    }
\caption{Examples of generated hypotheses from SciMON, \refinemethod, and findings from \cite{li2014generalrule}. Note that the SciMON idea is about creating a new method, where our hypothesis is about a new explanation for deception detection. We also show an existing finding from \citet{li2014generalrule} on deception detection, demonstrating that our generated hypothesis is highly relevant to the field of interest. }
\label{tab:comparison_with_scimon}
\end{table*}

\subsection{Example Hypotheses}
\label{appendix:hypothesis_examples:examples}
We include examples of generated hypotheses using our \unionhyporefine approach and \shortgpt , together with a brief qualitative analysis of its source in\begin{table*}[t]
    \centering
    \resizebox{1\textwidth}{!}{%
    \small
    \begin{tabular}{@{}lp{7cm}p{4.25cm}@{}}
    \toprule
    \textbf{Dataset} & \textbf{Generated Hypothesis} & \textbf{Literature Source/Novel} \\ \midrule \midrule
    \deceptive & Deceptive reviews often contain a higher frequency of first-person singular pronouns, while truthful reviews may use these pronouns less frequently. & \citet{li2014generalrule} \\ \\
    & The use of repetitive phrasing across multiple reviews is a strong indicator of deception, while truthful reviews are more likely to exhibit unique language and perspectives. & \citet{10.1007/s10489-022-03427-1} \\ \\
    & Reviews that provide specific accounts of the check-in and check-out processes, including exact times, the names of staff members involved, and descriptions of any unique features or services utilized (e.g., "I used the self-check-in kiosk at 3 PM"), are more likely to be truthful. Conversely, reviews that mention issues like long wait times or check-in problems without contextual details or specific examples (e.g., "the check-in took too long") are more likely to be deceptive. & Novel (from data) \\
    \midrule
    \gptgc and \llamagc & AI-generated content may struggle with maintaining coherence over longer passages, while human writing typically maintains clarity and focus. & \citet{tang2023sciencedetectingllmgeneratedtexts}\\ \\
    & AI-generated texts are more likely to follow conventional narrative structures, while human-written texts may experiment with form and structure. & Novel (from data)\\
    \midrule
    \dreaddit & Posts that show erratic posting behavior or changes in tone (e.g., from positive to negative) are more likely to indicate stress, while consistent posting patterns with a stable tone are more likely to indicate no stress. & \citet{userstressdetectionwan2024} \\ \\
    & Posts that exhibit avoidance behaviors (e.g., avoiding social situations or responsibilities) are more likely to indicate stress, while posts that demonstrate proactive engagement with challenges are more likely to indicate no stress. & \citet{Doan_2017}\\ \\
    & Posts that reflect on personal struggles with mental health or addiction (e.g., "I was a severe addict") are more likely to indicate that the poster has stress, while posts that discuss academic or professional experiences without emotional turmoil (e.g., "I've explained the aforementioned to people") are more likely to indicate that the poster does not have stress. & Novel (from data)\\
    \midrule
    \persuasion & Persuasive texts that incorporate rhetorical devices, such as rhetorical questions and direct appeals, are more likely to engage the reader and compel them to consider the writer's viewpoint. & \citet{WAGEMANS2023117} \\ \\
    & Texts that utilize strong, action-oriented verbs are generally more persuasive, as they convey confidence and urgency, compelling the audience to take action. & Novel (from data)\\ \\
    & Arguments that include a clear and compelling call to action are more persuasive, as they provide the audience with a specific next step to take, reinforcing the urgency and importance of the message. & Novel (from data)\\
    \bottomrule
    \end{tabular}
    }
\caption{Examples of generated hypotheses using our method accompanied by labels indicating their sources. For hypotheses from literature, we include the specific paper, while for hypotheses that are not explicitly suggested by our literature base, we set the label to "Novel (from data)".}
\label{tab:hypotheses-examples}
\end{table*}
\cref{tab:hypotheses-examples}. 
We also showcase example hypotheses generated using \notebooklm and \hyperwrite on \deceptive that are invalid or irrelevant in \cref{tab:invalid_examples}. These hypotheses can lead to degraded inference performance for theses two methods.
\begin{table*}[t]
    \centering
    \resizebox{1\textwidth}{!}{%
    \tiny
    \begin{tabular}{@{}lp{7cm}p{4.25cm}@{}}
    \toprule
    \textbf{Method} & \textbf{Invalid or Irrelevant Hypothesis} \\ \midrule \midrule
    \notebooklm & **Truthful reviews are more likely to be written in a style and tone that aligns with the reviewer's demographic information available on the platform, if any.** Conversely, deceptive reviews might exhibit inconsistencies between the writing style and the reviewer's claimed demographic, signaling a potential fabrication. \\ \\
    & **Truthful reviews are more likely to be posted at various times and days, reflecting the organic behavior of genuine guests.** Conversely, deceptive reviews, particularly those orchestrated by paid posters, might be posted in clusters or at unusual times, indicating a coordinated effort. \\ \\
    & **Truthful reviews are more likely to be written in a way that aligns with the overall sentiment expressed in the review's star rating.** Conversely, deceptive reviews might show inconsistency between the sentiment expressed in the written content and the assigned star rating, indicating a potential attempt to manipulate perception. \\
    \midrule
    \hyperwrite & **Relevant Images:** Truthful reviews are more likely to include relevant images. Deceptive reviews less likely to include images. \\ \\
    & **First-Person Pronouns:** Truthful reviews use first-person pronouns (I, my). Deceptive reviews use third-person (one). \\ \\
    & **Overly Formal Language:** Deceptive reviews use overly formal language. Truthful reviews use conversational tone. \\
    \bottomrule
    \end{tabular}
    }
\caption{Examples of generated hypotheses using \notebooklm and \hyperwrite on \deceptive that are invalid or irrelevant, leading to degraded inference performance for these methods.}
\label{tab:invalid_examples}
\end{table*}

\end{document}